\documentclass[runningheads]{llncs}

\usepackage{eccv}

\usepackage{eccvabbrv}

\usepackage{graphicx}
\usepackage{booktabs}

\usepackage[accsupp]{axessibility}  
\usepackage{subcaption}
\usepackage{color}
\usepackage{colortbl} %
\usepackage{tabularx}
\usepackage[table]{xcolor}

\usepackage{amsmath}  
\usepackage{amssymb}

\definecolor{tableblue}{HTML}{E6F2FF}
\definecolor{tablegrey}{HTML}{F2F2F2}
\definecolor{gaincolor}{HTML}{008080}

\newcolumntype{Y}{>{\centering\arraybackslash}X}

\usepackage{hyperref}
\newcommand{\equalcontrib}{\textsuperscript{*}}
\usepackage{orcidlink}
\usepackage{wrapfig}
\usepackage{caption}
\usepackage{algorithm}
\usepackage{algpseudocode} 

\usepackage[most]{tcolorbox}

\newtcolorbox[blend into=figures]{promptbox}[2][]{%
    colback=gray!5,
    colframe=gray!30,
    sharp corners,
    boxrule=0.5pt,
    left=10pt, right=10pt, top=10pt, bottom=10pt,
    fontupper=\small\ttfamily\raggedright,
    breakable,
    before upper={\setlength{\parindent}{0pt}},
    title={#2},              
    coltitle=black,          
    colbacktitle=gray!15,    
    fonttitle=\bfseries\small,
    every float=\centering,  
    #1                       
}

\author{
  Xuanzhao Dong\inst{1}\thanks{These authors contributed equally to this work.} \and
  Wenhui Zhu\inst{1,\equalcontrib} \and
  Peijie Qiu\inst{3,\equalcontrib} \and
  Xiwen Chen\inst{2,\equalcontrib} \and
  Xiaobing Yu\inst{3} \and
  Xin Li\inst{1} \and
  Zhipeng Wang\inst{6} \and
  Shao Tang\inst{5} \and
  Gen Li\inst{3} \and
  Yujian Xiong\inst{1} \and
  Hao Wang\inst{2} \and
  Yanxi Chen\inst{1} \and
  Prayag Tiwari\inst{4} \and
  Yalin Wang\inst{1}\thanks{Corresponding author: \texttt{ylwang@asu.edu}}
  \protect\protect\footnotetext{Preprint.}
}

\institute{
  Arizona State University, Tempe, AZ, USA \and
  Clemson University, Clemson, SC, USA \and
  Washington University in St. Louis, St. Louis, MO, USA \and
  Halmstad University, Halmstad, Sweden \and
  Florida State University, Tallahassee, FL, USA \and
  Rice University, Houston, TX, USA
}

\title{Mags-RL: Wearing Multimodal LLMs a Magnifying Glass via Agentic Reinforcement Learning For Complex Scene Reasoning} 

\titlerunning{Mags-RL: Wearing Multimodal LLMs a Magnifying Glass}

\authorrunning{X. Dong et al.}

\begin{document}
\maketitle

\begin{abstract}
Despite their popularity and success, Multimodal Large Language Models (MLLMs) often struggle to interpret images accurately, which limits their reasoning capability in complex scenarios (e.g., high object density and complex background clutter). Prior work mainly addresses this limitation by incorporating explicit visual cues like bounding boxes that require extra annotations. In addition, the resulting low-resolution crops often miss fine-grained details that MLLMs require for accurate reasoning.
Therefore, we propose \textbf{Mags-RL}, an Agentic Reinforcement Learning (RL) framework that wears MLLMs with an external super-resolution “magnifying glass” agent for high-resolution fine-grained inspection. Specifically, the model performs two-round reasoning: in the first round, it generates an initial rationale and autonomously identifies regions of interest without relying on additional annotations; in the second round, it invokes a super-resolution agent to crop and upscale those regions, then revisits and verifies its earlier reasoning to produce the final answer. We also introduce a novel curriculum learning strategy that enables a data-efficient RL training, needing as few as \textbf{only 40 training samples} to achieve reasonable performance. Experiments on VSR, TallyQA, and GQA subsets show its superior performance against recent strong competing methods, demonstrating high-quality reasoning with precise visual grounding. Code and weights will be released soon.

\end{abstract}

\section{Introduction}\label{sec:intro}
The introduction of Multimodal Large Language Models (MLLMs) has significantly altered the course of vision-language research~\cite{achiam2023gpt, team2023gemini, chen2025aha, chen2026prompt, qiu2025multimodal}. By integrating textual and visual modalities, these models have demonstrated remarkable progress across diverse domains~\cite{yin2024survey,wu2023multimodal,caffagni2024revolution}. While early iterations excelled at basic visual recognition and cross-modal alignment, solving complex real-world tasks requires human-level perceptions. Consequently, the research frontier has naturally shifted towards deep logic understanding, driving a heavy focus on enhancing model reasoning capabilities~\cite{zhou2025perception,wang2024exploring, li2025perception,chen2025dra}. This allows the model to expand beyond purely text-driven logic and incorporate visual context directly into their reasoning chains~\cite{su2025thinking,yang2025deep,shao2024visual}. 

\begin{figure}[!t]
    \centering
    \includegraphics[width=1.0\textwidth]{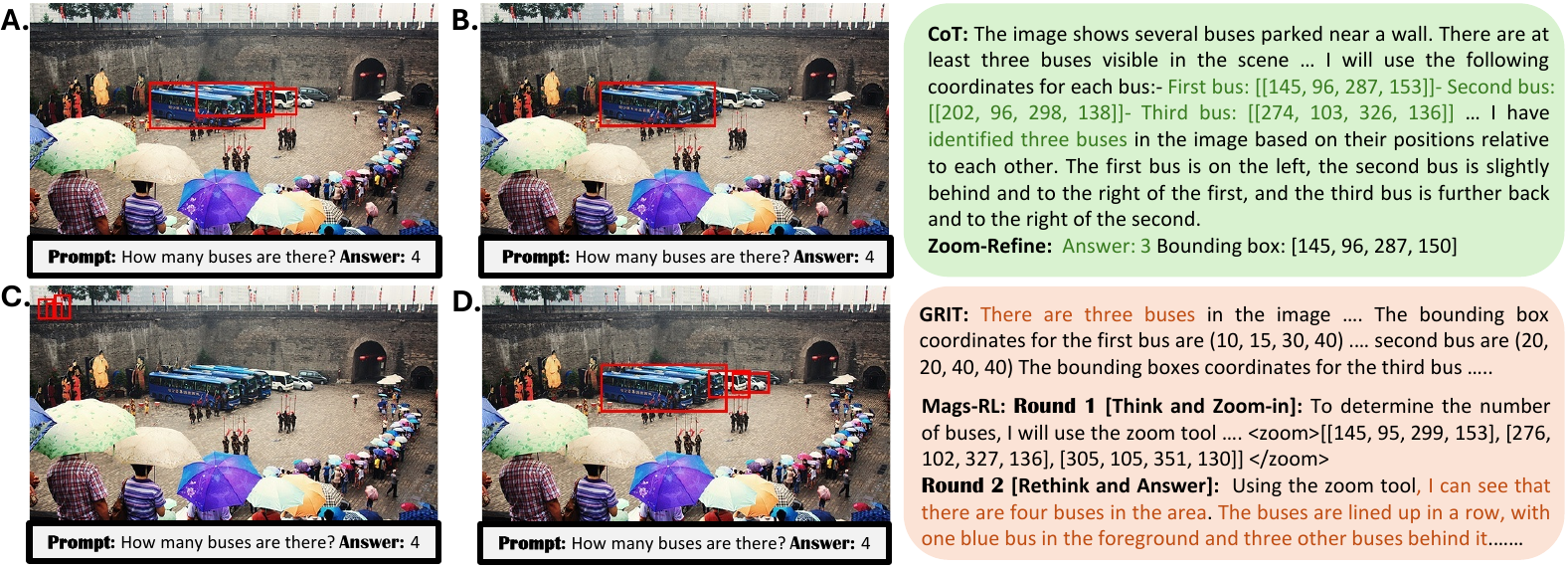} 
    \caption{Illustration of model responses to counting questions from TallyQA. \textbf{A.}-\textbf{D.} show the responses produced by CoT, Zoom-Refine, GRIT, and Ours (i.e., Mags-RL), respectively. The \textbf{Red} boxes denote the image crops generated from the model-predicted coordinates. Key syntax is omitted for clarity.}
    \label{fig:first-exp}
\end{figure}

Among these efforts, prompt-based methods represent a prominent solution and have shown promising results~\cite{wei2022chain,zhang2023multimodal}. As shown in Fig.~\ref{fig:first-exp}\textbf{A}, by carefully designing the expected reasoning behavior, models can generate accurate reasoning chains that interact with visual content through bounding box coordinates. However, their sensitivity to prompt design continues to limit performance in complex scenarios~\cite{schmalfuss2025parc}. As shown in Fig.~\ref{fig:first-exp}\textbf{B}, when the system prompt explicitly requires the generation of only the most relevant bounding box, the model correctly recognizes major components (e.g., the blue buses) but overlooks peripheral yet relevant objects, such as a partially occluded white bus in the background. 

In contrast to the aforementioned training-free approaches, some methods propose interleaving reasoning with image-grounded bounding box generation within textual responses~\cite{peng2023kosmos,chen2023shikra,fan2025grit,wu2025grounded}, which typically leads to response inconsistency. As illustrated in Fig.~\ref{fig:first-exp}\textbf{C}, the generated coordinates drift toward the top-left corner of the guardrail rather than the bus location described in the text. In this case, the bounding box output behaves more like a trivial token sequence than a meaningful intermediate step that contributes to a correct response. These failures underscore a critical need for robust reasoning paradigms that transcend fragile prompt engineering. Genuine "thinking with images" demands dynamic interaction with the visual content itself, rather than relying on the pseudo-communication of purely textual bounding box sequences~\cite{wu2025reinforcing}. Motivated by human visual cognition, which typically establishes a global context before zooming in on specific areas of interest, we argue that \emph{empowering models to actively isolate and scrutinize fine-grained visual details will effectively overcome these reasoning bottlenecks in complex scenarios.}


To this end, we propose Mags-RL, a novel agentic reinforcement learning framework that is explicitly designed to improve the reasoning abilities of MLLMs in complex scenarios by equipping the model with a virtual "magnifying glass." Unlike pure text coordinate generation, Mags-RL enables "thinking with images" through a two-round reasoning process. Specifically, in the first round, the model generates an initial rationale and autonomously identifies regions of interest without relying on additional annotations. It then interacts with an external super-resolution agent designed to crop and upscale those specific areas. Finally, upon receiving this enriched visual feedback, the model revises its initial logic to deduce the final answer. As shown in Fig.~\ref{fig:first-exp}\textbf{D}, by clearly identifying three potential regions containing buses, this two-round interaction makes the model successfully identify all four buses along with their spatial relationships (e.g., noting the blue bus in the front). Furthermore, to ensure system consistency and data efficiency, we train the model using the Group-Relative Policy Optimization (GRPO)~\cite{shao2024deepseekmath} algorithm with as few as 40 samples, guided by carefully curated format- and accuracy-based reward signals alongside curriculum learning (CL)~\cite{bengio2009curriculum} dynamics. To demonstrate the advantages of Mags-RL, we conduct extensive experiments across three major benchmarks (i.e., VSR~\cite{liu2023visual}, TallyQA~\cite{acharya2019tallyqa}, and GQA~\cite{hudson2019gqa}) encompassing three different levels of reasoning complexity, where Mags-RL achieves superior performance compared to baselines. Finally, we provide two in-depth analyses evaluating the super-resolution (SR) module against a direct crop-and-resize (CR) implementation, as well as the impact of the CL training strategy. 

In summary, our contributions are as follows:
\begin{itemize}
    \item We propose Mags-RL, a novel agentic reinforcement learning framework that improves the reasoning abilities of MLLMs in complex visual scenarios by a two-round reasoning process through dynamic interaction with an external super-resolution agent. 
    
    \item We introduce a highly data-efficient training paradigm for multimodal reasoning built upon GRPO algorithms. With our curated reward signals and curriculum learning dynamics, we successfully align the model's reasoning behaviors using as few as 40 samples, drastically reducing data dependency.
    
    \item Through comprehensive evaluations across three major benchmarks encompassing varying levels of reasoning complexity, we demonstrate the superior performance Mags-RL against several contemporary baselines, showing its ability to shifts the MLLM's response behavior toward a more robust, agent-oriented reasoning pattern. 
    
\end{itemize}




\section{Related Work}

\noindent \textbf{Multimodal Large Language Models (MLLMs).} The rapid progress of MLLMs has altered the course of vision–language interaction. Open-source efforts have been instrumental in delivering strong models across domains such as medical imaging~\cite{zhu2025retinalgpt,dong2025llada,yang2024advancing, li2023llava}, autonomous driving~\cite{sima2024drivelm,xu2024drivegpt4,shao2024lmdrive}, and robotics~\cite{ahn2022can,driess2023palm,brohan2023rt2visionlanguageactionmodelstransfer}. Beyond these domain-specific applications, recent work has increasingly targeted on improving the reasoning ability of MLLMs. Early prompt-based approaches include In-Context Learning~\cite{brown2020languagemodelsfewshotlearners} (ICL) and Chain-of-Thought~\cite{wei2022chain} (CoT), which inject in-context examples or intermediate reasoning steps into the prompt to steer the model toward task-aligned outputs. Zoom-Refine~\cite{yu2025zoomrefineboostinghighresolutionmultimodal} goes beyond prompt engineering by introducing an explicit response-and-revise workflow, prompting the model to reevaluate its earlier rationale using additional visual evidence. In contrast, GRIT~\cite{fan2025grit} advocates grounded reasoning by producing reasoning chains that interleave natural language with explicit bounding-box coordinates.
Despite these advances, major limitations persist, including the brittleness of prompt-based methods and the underutilization of fine-grained visual interaction. As a result, existing vision–language reasoning paradigms still struggle in complex scenes, particularly those with dense object clustering or extreme viewpoints. To address these challenges, Mags-RL offers a more robust alternative: by enabling dynamic interaction with super-resolution image crops, it substantially strengthens fine-grained visual understanding and grounding in difficult scenarios.

\noindent \textbf{Agentic Reinforcement Learning.} Reinforcement learning from human feedback (RLHF)~\cite{christiano2017deep, ouyang2022training} have demonstrated strong potential for improving model reasoning and preference alignment. For example, MPO~\cite{wang2024enhancing} enhances multimodal reasoning by jointly optimizing preference signals and response quality, thereby mitigating issues such as repetitive or low-quality generations. However, these approaches are typically studied in relatively static settings, where the target output is well defined (e.g., mathematical reasoning or question answering). As the focus shifts from relatively static prediction tasks to dynamic interaction with external environments, the limitations of conventional single-turn post-training methods become more pronounced~\cite{zhang2025generalizability, chen2025ai}. In such settings, agentic reinforcement learning (RL) frameworks~\cite{zhang2025landscape, cheng2025agent}, which train models to manage system interactions, invoke tools, and execute sequential actions, have emerged as a promising complement. Recent work has shown the potential of this paradigm. For instance, LaMer~\cite{jiang2025meta} introduces a cross-episode training framework with in-context policy adaptation, enabling agents to explore effectively and improve from test-time feedback. Nevertheless, the integration of agentic RL into multimodal architectures, particularly in complex scenarios requiring deep visual comprehension beyond superficial cues, remains a largely underexplored frontier. To address this gap, we introduce Mags-RL, a data-efficient RL paradigm that enhances reasoning by facilitating interaction with external visual tools. The underlying motivation is highly intuitive. By provisioning the model with an external "magnifying glass," Mags-RL empowers the agent to dynamically zoom in and analyze fine-grained details, driven entirely by its intrinsic reasoning objectives.

\begin{figure}[!t]
    \centering
    \includegraphics[width=0.85\textwidth]{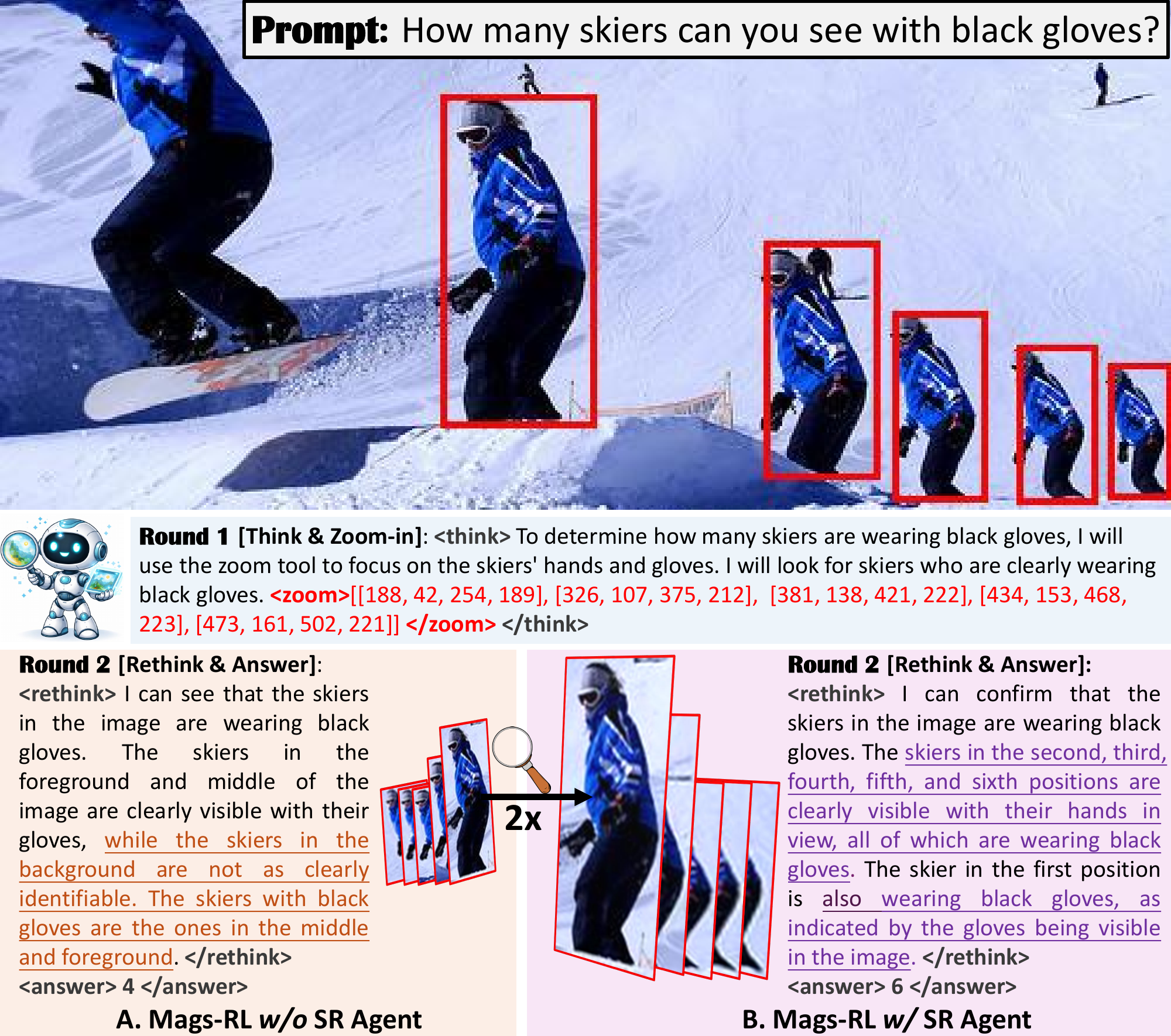} 
    \caption{Overview of the Mags-RL pipeline, which is comprised of a two-round reasoning chain: i) the LLM policy generates an initial rationale and determines a zoom-in region based on the multimodal input; and ii) the LLM policy performs an additional reasoning step to verify the initial logic and generate the final answer. In between these two rounds, a magnifying super-resolution (SR) agent is employed. Mags-RL equipped with the SR agent (\textbf{B}) exhibits stronger reasoning capabilities than the one without it (\textbf{A}).}
    \label{fig:mags-method}
\end{figure}

\section{Methods}
This section details the proposed Mags-RL framework. We start by introducing the rule-based RL training pipeline in Sec.~\ref{subsec:preli}. Sec.~\ref{subsec:mags_sample} formulates our two-round agentic reasoning. Sec.~\ref{subsec:reward-obj} introduces our reward design. Sec.~\ref{subsec:curriculum-learning} presents our novel curriculum learning strategy.

\subsection{Preliminaries}\label{subsec:preli}
\noindent \textbf{Multimodal  Large Language Models (MLLMs).} MLLMs typically modeled as autoregressive text generators that can process multiple modalities. In this work, we focus on the setting where the input consists of an image $\mathbf{I}$ and a text question $\mathbf{q}$. Formally, given an input $\mathbf{x} = [\mathbf{I}, \mathbf{q}]$, the MLLM $\pi_\theta(\cdot |\mathbf{x})$ generates the output token sequence $\mathbf{o} = \{o_1, \cdots, o_t, \cdots, o_T\}$. 

\noindent \textbf{GRPO for pos-training MLLMs.} Reinforcement learning from human feedback has emerged as the de facto standard for training reasoning models. However, traditional Proximal Policy Optimization (PPO)~\cite{schulman2017proximal} is cumbersome due to its reliance on separate reward and value models. With the introduction of Group Relative Policy Optimization (GRPO)~\cite{shao2024deepseekmath}, rule-based rewards offer a more efficient alternative, delivering strong empirical performance for training reasoning models. Specifically, the objective GRPO is formulated as:
\begin{equation}\label{eq:grpo}\nonumber
    \mathcal{L}_{\text{GRPO}}(\theta) = \mathbb{E} \left\{ \frac{1}{G} \sum_{i=1}^{G} \frac{1}{|o_i|} \sum_{t=1}^{|o_i|} \left[ \min\left(s_{i, t} \hat{A}_{i, t}, \text{clip}(s_{i,t}, 1 - \epsilon, 1 + \epsilon) \hat{A}_{i, t} \right) \right] \right\},
\end{equation}
where the importance sampling ratio between new and old LLM policy is defined as $s_{i, t} = \frac{\pi_\theta(o_{i, t} | \mathbf{x}, o_{i, <t})}{\pi_{\text{old}}(o_{i, t} | \mathbf{x}, o_{i, <t})}$, $\hat{A}_{i,t}$ is the advantage calculated based on relative rewards of the outputs inside each group, and $\epsilon$ is a hyper-parameter. In practice, there is a KL divergence regularization $\mathbb{D}_{\text{KL}}[\pi_\theta || \pi_{\text{ref}}]$, is applied between the trained policy and a reference policy (typically a supervised fine-tuned (SFT) model). This term is scaled by a weight parameter $\beta$ to ensure the trained policy remains close to the reference one. For notational simplicity, we omit the KL term hereafter.


\subsection{Two-Round Agentic Reasoning via Magnification}\label{subsec:mags_sample}
Unlike single-pass reasoning in standard MLLMs, we decompose the reasoning path into two rounds. This design is largely motivated by human perception: we first form a global overview, then narrow our attention to task-relevant details by further analyzing those regions of interest. To this end, our two-round reasoning path is as follows (see Fig.~\ref{fig:mags-method}): 
\begin{itemize}
    \item \textbf{Round 1}: The model identifies a specific region of interest by generating output bounding boxes inside a special $\texttt{<zoom>}$ tag;
    \item \textbf{Round 2}: The model revisits the zoomed-in regions along with the initial reasoning rationale and generates the final answers.
\end{itemize}
\noindent Formally, the \textbf{Round 1} output comprises an initial reasoning trace $\mathbf{o}_{\text{think}}^{(1)}$ enclosed by the \texttt{<think>} tag and a zoom specification $\mathbf{o}_{\text{zoom}}^{(1)}$ enclosed by the \texttt{<zoom>} tag. Notably, unlike prior work, our model does not rely on any additional grounding information.  In \textbf{Round 2}, the model conditions on the original question $\mathbf{q}$, the original image $\mathbf{I}$, the first-round reasoning trace $\mathbf{o}_{\text{think}}^{(1)}$, and the proposed zoomed-in region $\mathbf{o}_{\text{zoom}}^{(1)}$, and then generates a refined reasoning trajectory $\mathbf{o}_{\text{think}}^{(2)}$ along with the final answers $\mathbf{a}$.

However, \emph{how to inject the zoomed-in crops into MLLMs remains a critical yet underexplored question.} A naive solution is to treat the $k$ zoomed-in crops $\mathbf{Z}:= [\mathbf{z}_1, \cdots, \mathbf{z}_k]$ as additional visual prompts~\cite{li2024monkey}. However, our initial investigation showed that this approach performs poorly, largely because MLLMs struggle to reason over low-resolution crops, particularly when the zoomed-in crops are small (see Fig.~\ref{fig:mags-method}\textbf{A}). To address this limitation, we leverage an external super-resolution (SR) agent $S_a$ to “magnify” the zoomed-in crops, $\mathbf{Z}^{\text{SR}}:= [\mathbf{z}_1^{\text{SR}}, \cdots, \mathbf{z}_k^{\text{SR}}]$, enabling the model to reason over high-resolution crops and capture finer-grained details (see Fig.~\ref{fig:mags-method}\textbf{B}), particularly for complex tasks such as counting. 

\subsection{Reward Design for Agentic RL}\label{subsec:reward-obj}
To encourage effective reasoning and correct interaction with the external agent, we tailor the reward design accordingly. Specifically, we introduce three reward signals: (i) a format reward $r_{\text{fmt}}$ that enforces a strict sequential structure in the model’s responses, thereby preserving the logical flow of the agentic loop; (ii) an answer-accuracy reward $r_\text{ans}$, which provides the primary supervision to ensure the final output is both concise and correct; and (iii) a zoom-accuracy reward $r_\text{zoom}$, which incentivizes $\pi_\theta$ to produce plausible, well-grounded coordinate predictions consistent with its initial reasoning. We outline the core designs below and direct readers to \textbf{Appendix B} for additional details on the reward design.

\begin{figure}[!t]
    \centering
    \includegraphics[width=0.85\textwidth, height=0.2\textheight]{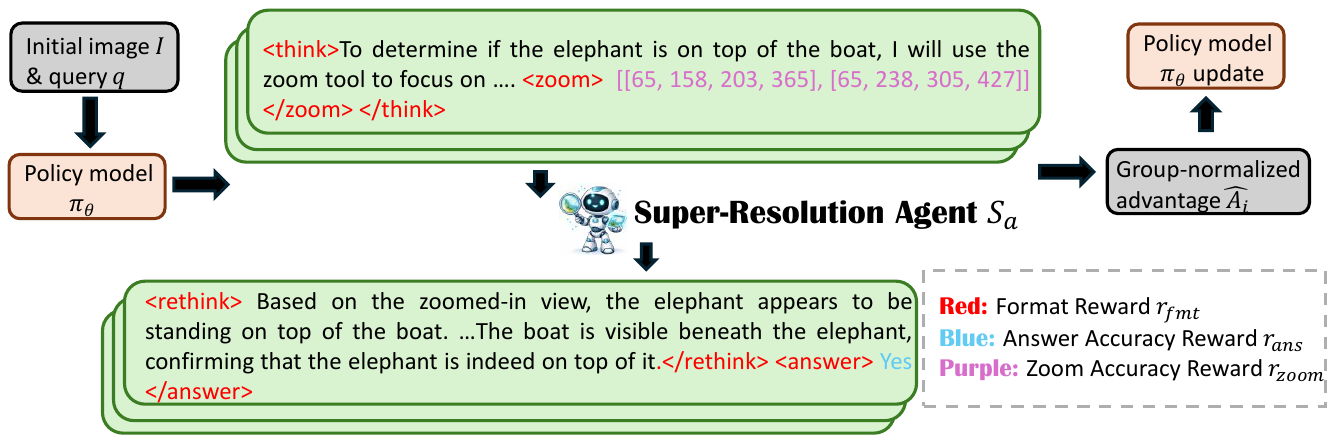}
    \caption{\textbf{Illustration of the Reward Design for Mags-RL.} The final reward is a sum of all reward signals. See Sec.~\ref{subsec:reward-obj} for more details. }
    \label{fig:mags-reward}
    \vspace{-0.13cm}
\end{figure}

\noindent $\bullet$ \textbf{Format Reward} $r_{\text{fmt}}$. This reward incentivizes the policy model $\pi_\theta$ to generate output sequences that strictly adhere to predefined structural constraints. Specifically, we assign a reward of 1 point to each of the required tags \texttt{<zoom>} and \texttt{<answer>}, and 0.5 point to each of the required tags \texttt{<think>} and \texttt{<rethink>}. This yields a maximum format reward of 3 (see Fig.~\ref{fig:mags-reward}; red highlights). If any required tag is missing, the total reward is reduced by the corresponding amount assigned to that tag.

\noindent $\bullet$ \textbf{Answer Accuracy Reward} $r_{\text{ans}}$. This reward serves as the primary signal to ensure the correctness of the answer. Specifically, we formulate this reward as a hierarchical signal. As shown in Fig.~\ref{fig:mags-reward}, we only give a full point of 1 once answer $\mathbf{a}$ is an exact match of the ground-truth $\mathbf{a}^*$. Additionally, to handle cases where semantic meaning is correct despite a surface-level mismatch (e.g., "cat" vs. "kitty"), we give partial points if the similarity score generated by the LLM judge based on the triplet $(\mathbf{q},\mathbf{a}^*,\mathbf{a})$ is higher enough. 

\noindent $\bullet$  \textbf{Zoom Accuracy Reward} $r_{\text{zoom}}$. This reward encourages the LLM policy to generate precise bounding boxes, enabling the $S_a$ to accurately crop and upscale the relevant regions of interest. The reward signal is structured according to two hierarchical curriculum learning stages (See.~\ref{subsec:curriculum-learning}). In the initial stage, we prioritize the precision of valid bounding boxes (see Fig.~\ref{fig:mags-reward}; purple highlights) to encourage correct yet conservative predictions. In the second stage, particularly for counting-based tasks, the objective shifts from precision to recall to enforce higher coverage and more exhaustive spatial exploration. Notably, our reward design does not rely on ground-truth bounding box supervision, as such labels are often inconsistent in complex scenarios (e.g., cases where only two sets of coordinates are provided for more than seven targets).

\begin{figure}[!t]
    \centering
    \includegraphics[width=0.95\textwidth]{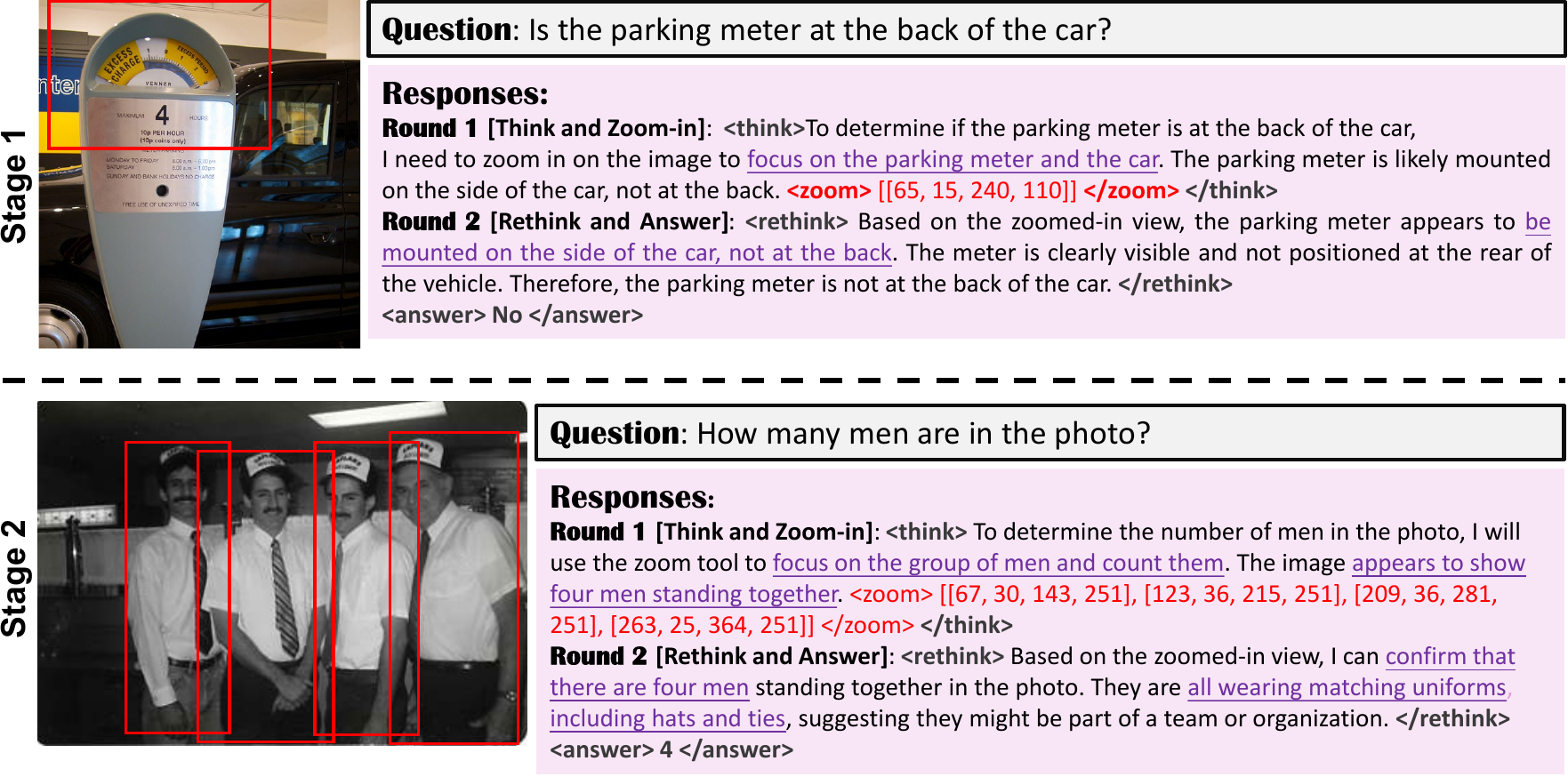}
    \caption{Comparison of generation results across curriculum learning stages. In stage 1, the model successfully acquires the ability to perform reasoning, trigger the external SR agent, and execute self-verification. However, its bounding box predictions remain relatively conservative, often under-covering relevant targets. In stage 2, the model not only maintains structural correctness but also demonstrates improved spatial reasoning, generating bounding boxes that accurately encompass the regions of interest, which is a prerequisite for fine-grained image understanding in counting tasks. }
    \label{fig:mags-curriculum}
\end{figure}

\subsection{Curriculum Learning for Mags-RL}\label{subsec:curri-grpo}
Simultaneously training the LLM policy $\pi_\theta$ to adhere to syntax requirements, identify key crops under weak supervisory signal (i.e., without using ground-truth bounding boxes), and provide accurate answers results in severe training instability due to the high optimization complexity (see Sec.~\ref{subsec:curriculum-learning} for our ablation study). To mitigate this, we implement a two-stage curriculum learning in the Mags-RL training loop. In the first stage, we utilize 20 foundational training examples. This stage serves to initialize basic competencies, encouraging the model to follow structural requirements, invoke the external SR agent with properly generated bounding boxes, and answer questions concisely and accurately. In Stage 1 (Fig.~\ref{fig:mags-curriculum}(top)), the model acquires the fundamental ability to zoom in on areas dictated by the reasoning content $w_1$ (e.g., the parking meter and the car). However, driven by the precision-based reward formulation, the model tends to make overly conservative bounding box predictions. 

To correct this "lazy" zooming behavior, the second stage introduces an additional 20 complex training instances. Concurrently, we adjust the system prompt, prompt suffix, and the reward function $r_{\text{zoom}}$ to ensure the behavior consistency. This adjustment penalizes conservative predictions by explicitly rewarding recall, strictly requiring the model to identify the exact number of necessary bounding boxes for counting questions. In stage 2 (Fig.~\ref{fig:mags-curriculum}(bottom)), the model actively generates more contextually relevant bounding boxes across diverse tasks. Rather than limiting itself to counting questions, it proactively attends to auxiliary regions when they provide helpful context (See case studies in \textbf{Appendix D}). Details regarding data selection and prompt modifications are provided in Sec.~\ref{subsec:dataset} and \textbf{Appendix A}, respectively.

\section{Experiments}
\subsection{Experimental Setups}\label{subsec:dataset}
In this section, we detail the datasets, evaluation metrics, and baselines. Due to space constraints, we direct the reader to \textbf{Appendices A and C} for a more comprehensive overview (e.g., the training implementation).

\noindent \textbf{Training Data.} Training Mags-RL requires two subsets of data, as discussed in Sec.~\ref{subsec:curri-grpo}. Specifically, in stage 1, 20 foundational samples are evenly sampled from the TallyQA~\cite{acharya2019tallyqa} and VSR~\cite{liu2023visual} benchmarks. In stage 2, an additional 20 complex samples are selected from the TallyQA and GQA~\cite{hudson2019gqa} benchmarks.

\noindent \textbf{Evaluation Benchmarks.} Throughout our experiments, we use subsets from the VSR, TallyQA, and GQA benchmarks to comprehensively evaluate the model's visual understanding abilities. Specifically, we stratify the evaluation data into three disjoint sets based on varying levels of difficulty, as determined by a VLM-as-a-judge framework (specifically, \texttt{Qwen3-VL-4B-Instruct}~\cite{bai2025qwen3}). This categorization yields the \texttt{Zoom-Easy} (standard spatial relations), \texttt{Zoom-Medium} (scale variations and occlusions), and \texttt{Zoom-Hard} (long-tail/dense objects
and extreme viewpoints) subsets. For evaluation, we employ GPT Accuracy (denoted as GPT) and Inclusion Accuracy (denoted as Inc.) as our primary metrics.

\noindent \textbf{Baselines.} we compare our approach against several recent strong baselines built upon the same base MLLM, including Direct Query, Chain-of-Thought (CoT)~\cite{wei2022chain}, One-shot In-Context Learning (ICL)~\cite{brown2020languagemodelsfewshotlearners}, GRIT~\cite{fan2025grit}, and Zoom-Refine~\cite{yu2025zoomrefineboostinghighresolutionmultimodal}. We use Qwen2.5-VL-3B-Instruct~\cite{bai2025qwen25vltechnicalreport} as our base model.

\subsection{Main Results}\label{subsec:analysis}

\begin{table*}[!t]
\centering
\caption{Quantitative results across different scene complexities. \textbf{GPT ACC} ($\uparrow$) and \textbf{Inc. ACC} ($\uparrow$) are reported in percentage (\%). The \textcolor{gaincolor}{$\blacktriangle$} denotes the absolute gain over the \textit{Direct Query} baseline. The best results of each column is highlighted in \textbf{bold}.}
\label{tab:main_results}

\small
\renewcommand{\arraystretch}{1.15} 
\setlength{\tabcolsep}{1pt}

\begin{tabularx}{0.9\textwidth}{l YYYYYY}
\toprule

\rowcolor{tablegrey} \textbf{ZOOM-Easy (Simple)} & \multicolumn{2}{c}{VSR} & \multicolumn{2}{c}{TallyQA} & \multicolumn{2}{c}{GQA} \\
\cmidrule(lr){2-3} \cmidrule(lr){4-5} \cmidrule(lr){6-7}
Method & GPT $\uparrow$ & Inc. $\uparrow$ & GPT $\uparrow$ & Inc. $\uparrow$ & GPT $\uparrow$ & Inc. $\uparrow$ \\
\midrule
Direct Query     & 54.34 & 53.13 & 37.73 & 35.85 & 62.25 & 48.00 \\
Chain-of-Thought & 61.55 & 59.38 & 43.23 & 40.33 & 61.20 & 45.20 \\
One-shot ICL     & 63.02 & 60.07 & 31.67 & 30.96 & 62.30 & 42.60 \\
Zoom-Refine      & 68.14 & 66.67 & 45.01 & 42.97 & 62.90 & 45.60 \\
GRIT             & 55.56 & 47.92 & 44.61 & 42.85 & 61.45 & 45.60 \\
\rowcolor{tableblue} \textbf{Ours} & \textbf{73.09} {\scriptsize \textcolor{gaincolor}{$\blacktriangle$18.8}} & \textbf{70.14} {\scriptsize \textcolor{gaincolor}{$\blacktriangle$17.0}} & \textbf{51.27} {\scriptsize \textcolor{gaincolor}{$\blacktriangle$13.5}} & \textbf{51.12} {\scriptsize \textcolor{gaincolor}{$\blacktriangle$15.3}} & \textbf{68.75} {\scriptsize \textcolor{gaincolor}{$\blacktriangle$6.5}} & \textbf{50.40} {\scriptsize \textcolor{gaincolor}{$\blacktriangle$2.4}} \\

\addlinespace[0.6em] 

\rowcolor{tablegrey} \textbf{ZOOM-Medium (Moderate)} & \multicolumn{2}{c}{VSR} & \multicolumn{2}{c}{TallyQA} & \multicolumn{2}{c}{GQA} \\
\cmidrule(lr){2-3} \cmidrule(lr){4-5} \cmidrule(lr){6-7}
Direct Query     & 47.88 & 47.70 & 12.75 & 09.60 & 52.98 & 42.34 \\
Chain-of-Thought & 59.33 & 58.10 & 11.80 & 07.52 & 51.12 & 41.04 \\
One-shot ICL     & 57.55 & 54.80 & 10.20 & 08.40 & 50.20 & 37.34 \\
Zoom-Refine      & 65.05 & 63.10 & 18.15 & 15.80 & 54.43 & 41.44 \\
GRIT             & 54.40 & 46.60 & 15.05 & 14.20 & 50.55 & 38.04 \\
\rowcolor{tableblue} \textbf{Ours} & \textbf{67.63} {\scriptsize \textcolor{gaincolor}{$\blacktriangle$19.8}} & \textbf{64.10} {\scriptsize \textcolor{gaincolor}{$\blacktriangle$16.4}} & \textbf{24.35} {\scriptsize \textcolor{gaincolor}{$\blacktriangle$11.6}} & \textbf{24.20} {\scriptsize \textcolor{gaincolor}{$\blacktriangle$14.6}} & \textbf{59.31} {\scriptsize \textcolor{gaincolor}{$\blacktriangle$6.3}} & \textbf{45.15} {\scriptsize \textcolor{gaincolor}{$\blacktriangle$2.8}} \\

\addlinespace[0.6em]

\rowcolor{tablegrey} \textbf{ZOOM-Hard (Complex)} & \multicolumn{2}{c}{VSR} & \multicolumn{2}{c}{TallyQA} & \multicolumn{2}{c}{GQA} \\
\cmidrule(lr){2-3} \cmidrule(lr){4-5} \cmidrule(lr){6-7}
Direct Query     & 45.65 & 45.65 & 09.71 & 07.19 & 44.32 & 34.52 \\
Chain-of-Thought & 46.74 & 45.65 & 07.01 & 05.76 & 45.48 & 33.97 \\
One-shot ICL     & 45.11 & 39.13 & 05.22 & 03.60 & 48.56 & 34.79 \\
Zoom-Refine      & 52.72 & 50.00 & 10.61 & 07.91 & 46.78 & 35.06 \\
GRIT             & 45.11 & 30.43 & 10.61 & 09.35 & 43.63 & 29.86 \\
\rowcolor{tableblue} \textbf{Ours} & \textbf{55.98} {\scriptsize \textcolor{gaincolor}{$\blacktriangle$10.33}} & \textbf{50.00} {\scriptsize \textcolor{gaincolor}{$\blacktriangle$4.35}} & \textbf{16.78} {\scriptsize \textcolor{gaincolor}{$\blacktriangle$7.07}} & \textbf{17.27} {\scriptsize \textcolor{gaincolor}{$\blacktriangle$10.08}} & \textbf{51.44} {\scriptsize \textcolor{gaincolor}{$\blacktriangle$7.12}} & \textbf{35.89} {\scriptsize \textcolor{gaincolor}{$\blacktriangle$1.37}} \\
\bottomrule
\end{tabularx}
\end{table*}

\begin{figure}[!t]
    \centering
    \includegraphics[width=0.9\textwidth]{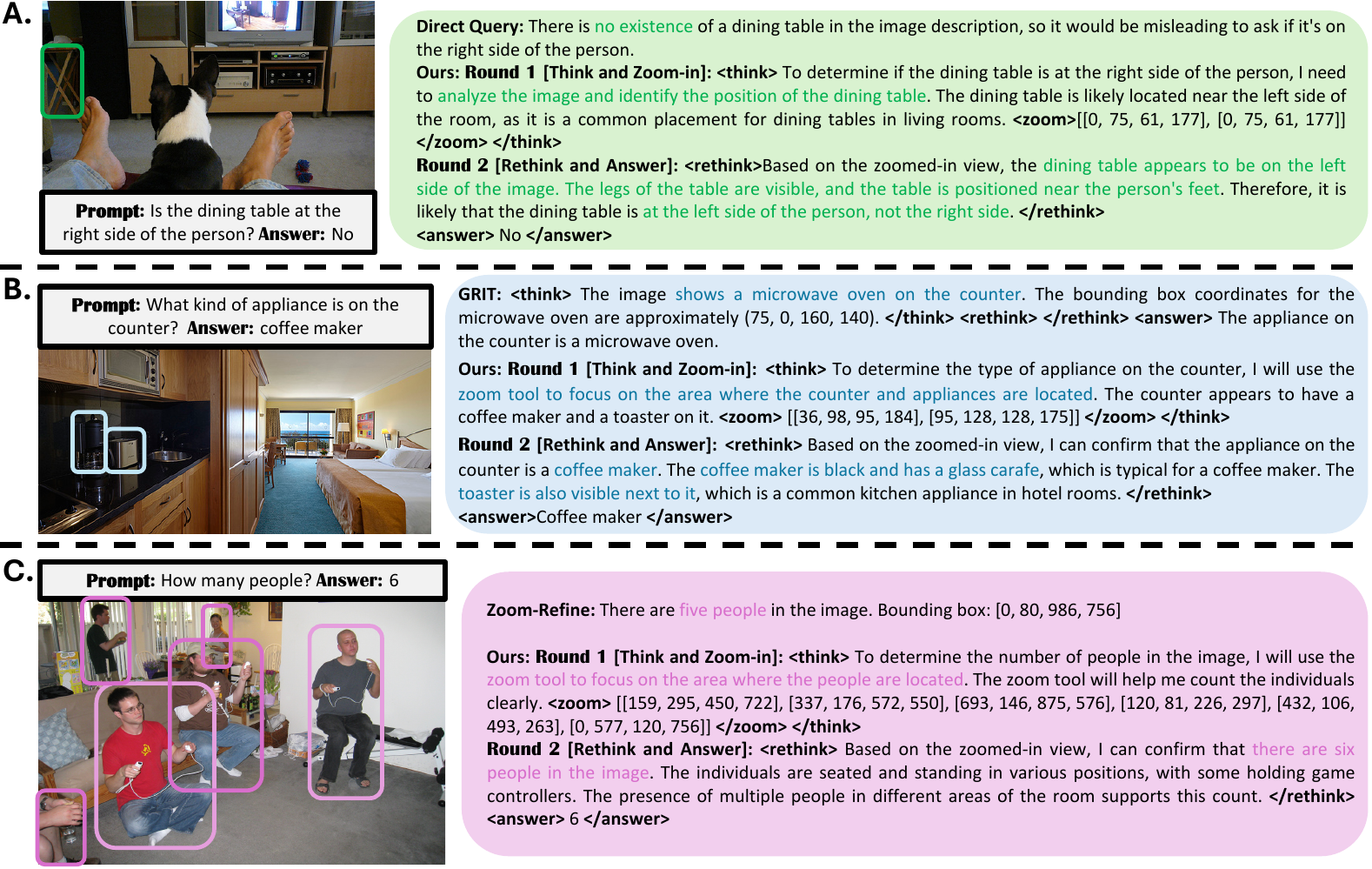}
    \caption{Illustrations of model responses from the \texttt{Zoom-Medium} benchmark. \textbf{A}. compares Mags-RL against the direct query baseline on the VSR dataset. \textbf{B}. compares Mags-RL against the GRIT baseline on the GQA dataset, and \textbf{C}. compares our method against the Zoom-Refine baseline on the TallyQA dataset.}
    \label{fig:baselines}
\end{figure}

Leveraging its ability to reason over and self-verify cropped regions, Mags-RL achieves superior performance across all three benchmark difficulty levels, demonstrating robust visual understanding in complex scenarios (see Table~\ref{tab:main_results}). On VSR, it outperforms the base model by 18.8\%, and retains a strong 10.3\% gain even at the hard difficulty level. On the Zoom-Hard subset of TallyQA, where dense object distributions require more advanced visual parsing, Mags-RL yields an improvement of approximately 7\% improvement over the base model. In addition, Mags-RL outperforms purely prompt-based methods or pseudo-visual interaction methods based on bounding-box generation. Notably, Mags-RL also shows strong robustness to different complex levels. For example, as TallyQA shifts from the simple level to the moderate level, the GPT-based accuracy of GRIT drop from 44.61\% to 15\%. Similarly, Zoom-Refine, which maintains strong performance in spatial reasoning and visual grounding through self-verification, declines from 45.01\% to 18.15\%.

Beyond these quantitative gains, qualitative results further underscore Mags-RL’s architectural strengths. First, the rethink-and-answer mechanism enables dynamic verification of the model’s initial reasoning. For instance, in Fig.~\ref{fig:baselines}\textbf{A}, zooming into the identified dining-table region allows Mags-RL to validate its spatial hypothesis and extract finer grounded cues (e.g., the table’s proximity to the subject’s feet), leading to the correct answer. Second, interaction with high-resolution crops yields stronger spatial grounding than baselines such as GRIT. As shown in Fig.~\ref{fig:baselines}\textbf{B}, GRIT generates bounding boxes primarily to guide reasoning over the original uncropped image, which can induce spatial hallucinations (e.g., mislocalizing the microwave). In contrast, Mags-RL correctly identifies the corner coffee maker and toaster and accurately resolves their relative layout. Finally, Mags-RL excels on more challenging counting tasks that demand fine-grained visual attention. As shown in Fig.~\ref{fig:baselines}\textbf{C}, Zoom-Refine self-corrects by focusing on a single dominant region, causing it to miss subtle evidence. By explicitly rewarding bounding-box recall, Mags-RL instead recovers highly occluded instances, such as a person visible only through their hands.


\begin{figure}[!t]
    \centering
    \includegraphics[width=0.85\textwidth]{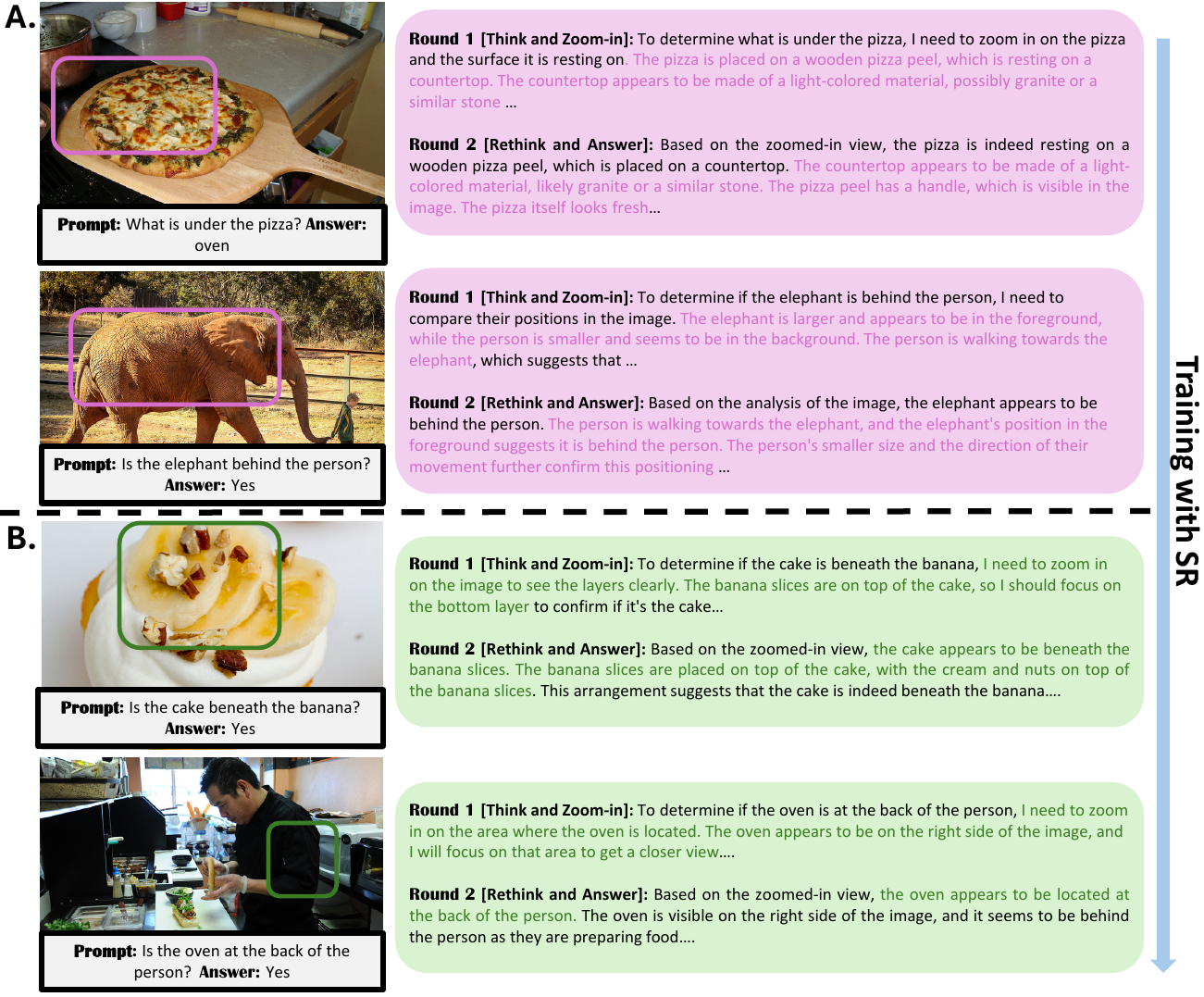}
    \caption{Illustration of model responses during CL stage-1. As training with Mags-RL progresses, the model gradually reduces overly descriptive outputs (sub-panel figures \textbf{A} and \textbf{B}) and instead tends to directly target the area of interest, thereby triggering the external visual agent. Key syntax is omitted for clarity. }
    \label{fig:train-with-sr}
\end{figure}

\subsection{The Effectiveness of the Super-Resolution Agent}\label{subsec:sr-influence}
When training with the super-resolution (SR) module, we observe an interesting shift in the model’s response behavior. Specifically, in the early stage of training (see Fig.~\ref{fig:train-with-sr}), the model devotes much of its reasoning chain to describing visual details in the image (e.g., the material of a countertop or the elephant and person). However, as training progresses, the model gradually adopts a more structured response pattern. As shown in Fig.~\ref{fig:train-with-sr}, the model first identifies the query-relevant region of interest (e.g., the bottom layer of banana slices or the location of the oven), and then performs rethinking and answer generation based on the returned system feedback.

\begin{figure}[!t]
    \centering
    \includegraphics[width=0.95\textwidth]{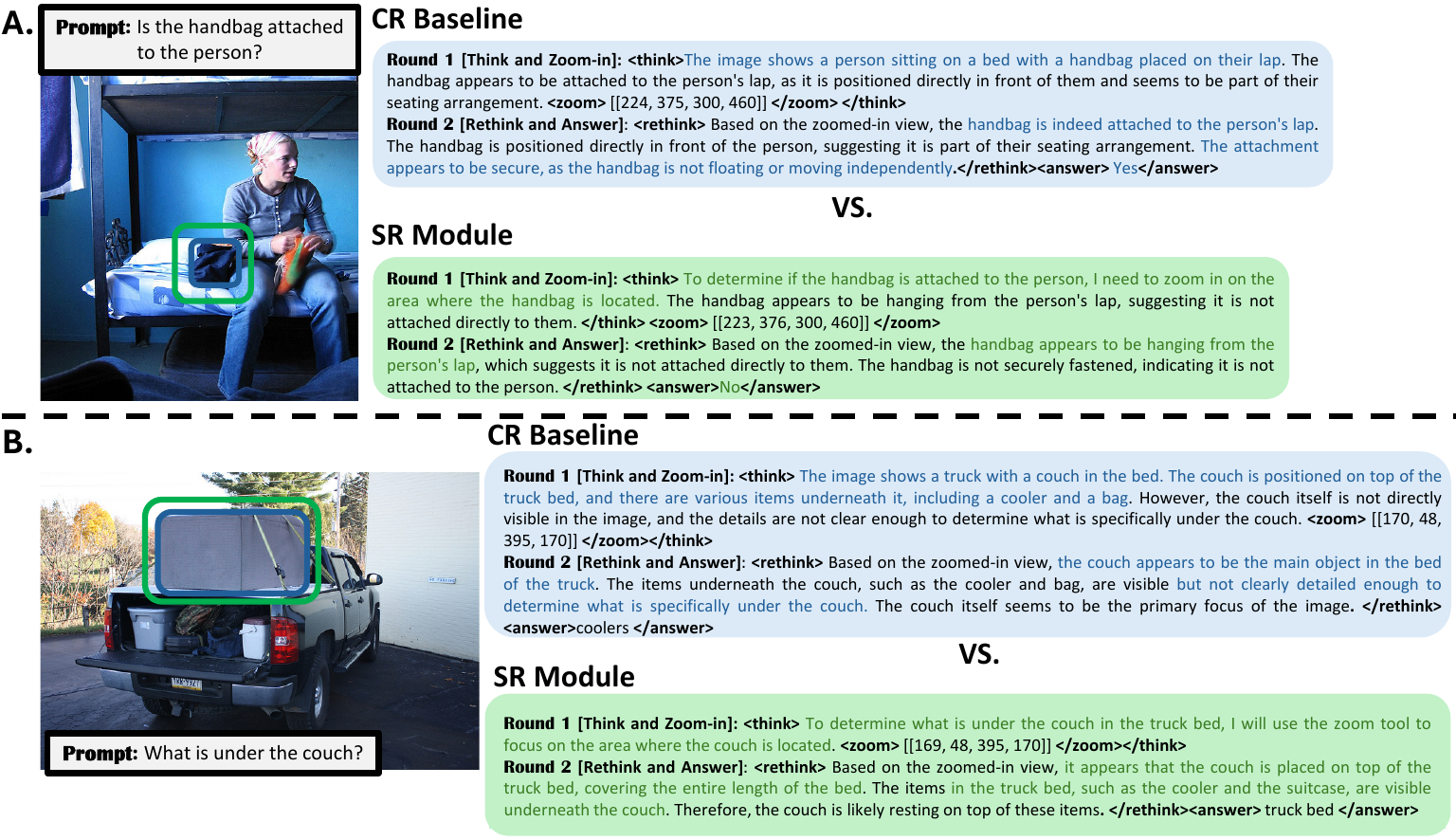}
    \caption{Comparison of model responses with and without the SR module. The \textbf{Blue} boxes highlight outputs from the Crop-and-Resize module, while the \textbf{Green} boxes indicate the response from our Stage-1 CL process. }
    \label{fig:example-sr-necessary}
\end{figure}

Motivated by this observed shift in response behavior, we further investigate the role of the SR agent in Mags-RL through an ablation study on its inclusion. Specifically, we use the CL Stage-1 dataset and keep all training settings unchanged, except that we replace the SR module with a direct cropping-and-resizing (CR) baseline. We find that the CR variant can still learn the desired syntax and successfully trigger the agent to invoke external visual components. For instance, as shown in Fig.~\ref{fig:example-sr-necessary}\textbf{A}, the CR baseline could describe image content (e.g., the person and handbag) and correctly identify and crop the relevant region using plausible coordinates. However, its intrinsic tendency to actively perform zoom-in reasoning is weakened, and the model instead reverts to the more descriptive behavior observed earlier in training (see Fig.~\ref{fig:train-with-sr}\textbf{A}). As shown in Fig.~\ref{fig:example-sr-necessary}\textbf{A}, although the CR baseline analyzes the image components (e.g., the cooler and the bag) and outputs the correct bounding box, it does so without explicitly connecting the zoom action to a targeted reasoning objective. This suggests that in the CR version, the correct coordinates are generated from the model's inherent knowledge priors, deriving from keywords in the text query (e.g., the couch) and visual encoder. In contrast, after incorporating the SR module, the model explicitly directs attention to the couch in order to determine what is underneath it. These findings suggest that the SR module may encourage a more agent-oriented response pattern, in which the model uses visual tool calls in a more purposeful and query-driven manner.

\begin{figure*}[!t]
    \centering
    \includegraphics[width=0.95\textwidth]{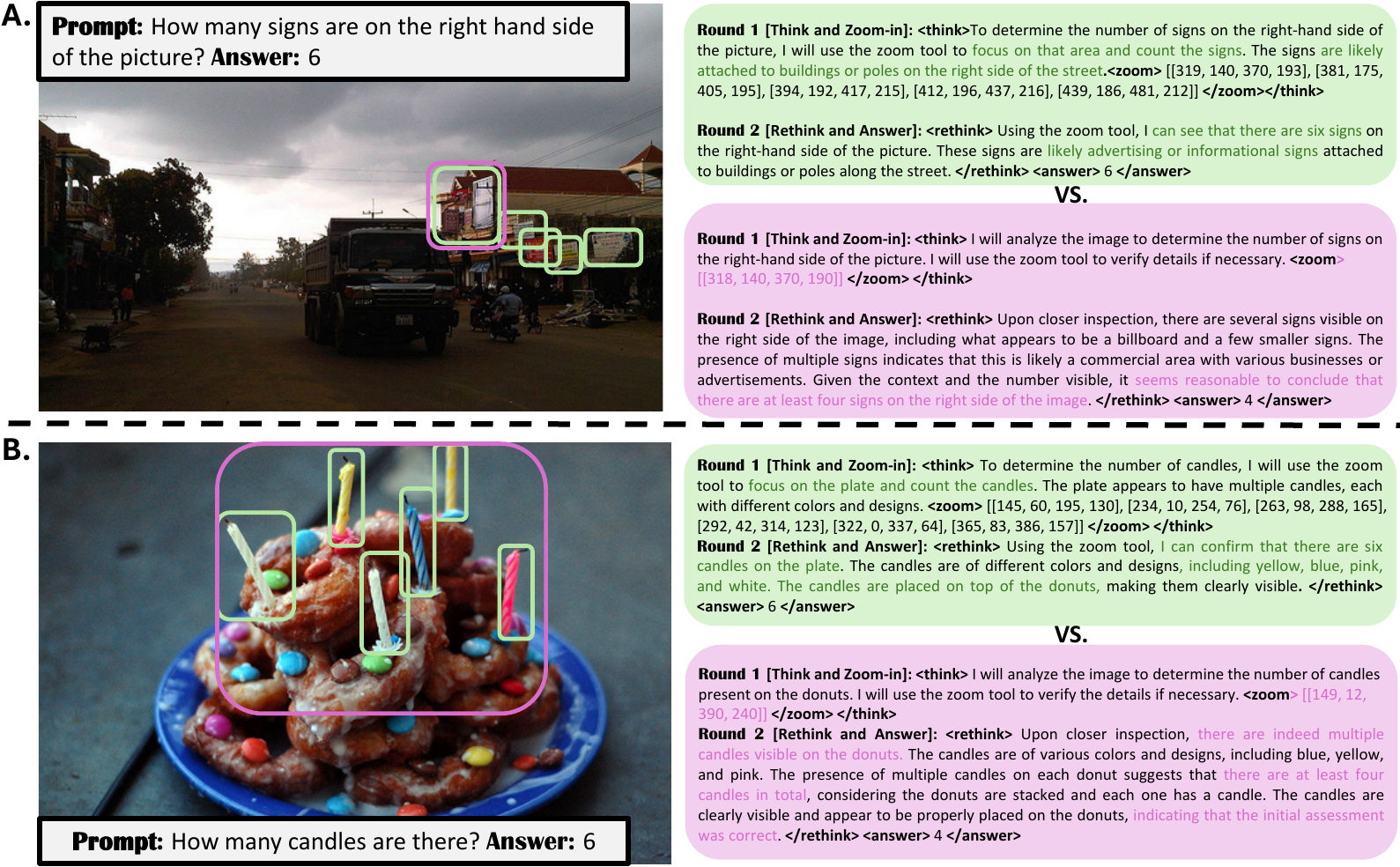}
    \caption{\textbf{Ablation study of the Curriculum Learning (CL) strategy}. The \textbf{green} box denotes results from CL training, while the \textbf{purple} box denotes results from Direct Training (DT) training. Although the DT model largely follows the required syntax and produces plausible content, it struggles to generate accurate bounding boxes for hard-level counting questions, resulting in more uncertain reasoning and final predictions, and thus more hallucinations. Colored highlights mark important text, and colored bounding boxes indicate crops from the corresponding model outputs.}
    \label{fig:cl-example}
\end{figure*}

\begin{table*}[t]
\centering
\caption{Ablation study of curriculum learning (CL) across three datasets in Zoom-Hard. See Sec.~\ref{subsec:curriculum-learning} for details.}
\scriptsize
\renewcommand{\arraystretch}{1.05}
\setlength{\tabcolsep}{4pt}

\begin{tabular*}{0.9\textwidth}{@{\extracolsep{\fill}} l cc cc cc @{}}
\toprule
\rowcolor{tablegrey}
\textbf{Zoom-Hard} 
& \multicolumn{2}{c}{\textbf{VSR}} 
& \multicolumn{2}{c}{\textbf{TallyQA}} 
& \multicolumn{2}{c}{\textbf{GQA}} \\
\cmidrule(lr){2-3} \cmidrule(lr){4-5} \cmidrule(lr){6-7}
& \textbf{GPT $\uparrow$} & \textbf{Inc. $\uparrow$}
& \textbf{GPT $\uparrow$} & \textbf{Inc. $\uparrow$}
& \textbf{GPT $\uparrow$} & \textbf{Inc. $\uparrow$} \\
\midrule
Direct training (DT)      & 53.80 & 47.83 & 12.41 & 12.23 & 46.58 & 30.41 \\
Curriculum learning (CL)  & 55.98 & 50.00 & 16.78 & 17.27 & 51.44 & 35.89 \\
\bottomrule
\end{tabular*}
\label{tab:ablation_CL}
\end{table*}

\subsection{The Impact of Curriculum Learning}\label{subsec:curriculum-learning}



Mags-RL trains the model $\pi_\theta$ using a two-stage curriculum learning (CL) strategy. To demonstrate the necessity of CL, we conduct an ablation study comparing CL with Direct Training (DT), and further evaluate both methods on the ZOOM-Hard benchmark. For a fair comparison, we keep the training configuration and datasets identical across settings, such that DT simulates training from scratch without curriculum scheduling. We find that the CL pipeline stabilizes the training dynamics, leading to better rethinking rationales and more accurate bounding box outputs, particularly in complex scenes with dense objects. As shown in Table~\ref{tab:ablation_CL}, training $\pi_\theta$ with CL improves both GPT score and inclusion accuracy across all three benchmarks. For example, it leads to around 4.37\% performance gain in TallyQA subset. This improvement is also reflected in more accurate bounding box predictions and more confident rethinking content. For example, in Fig.~\ref{fig:cl-example}\textbf{A}, compared with DT, the CL model correctly identifies all six small signs using five bounding boxes, whereas DT follows the desired output format but struggles to generate accurate coordinates. We argue that, without curriculum guidance, the model must optimize multiple reward signals simultaneously, especially generating appropriate bounding boxes without direct supervision while also maintaining answer correctness, which leads to a trade-off where zoom precision is sacrificed to preserve overall consistency. This behavior is further illustrated in Fig.~\ref{fig:cl-example}\textbf{B}: compared with the CL output, which produces separate crops for individual candles, DT instead generates one large bounding box covering all target objects in the image. As a result, the crop-based revision mechanism becomes less effective, and the model tends to draw conclusions from weak indications rather than explicit visual confirmation (e.g., “seems reasonable to conclude” or “indicating,” rather than “I can see” or “I confirm”).


\section{Conclusion}
In this paper, we present Mags-RL, a novel agentic reinforcement learning framework that equips MLLMs with a virtual "magnifying glass" to enhance visually grounded reasoning in complex scenarios. Through a two-round agentic generation process, the model first formulates a reasoning plan to identify critical regions of interest. It then invokes an external visual agent to crop and upscale these target areas based on the generated coordinates. Leveraging this enriched visual context, the model re-evaluates its initial rationale to deduce the final answer. Built upon the GRPO algorithm, Mags-RL demonstrates exceptional data efficiency, requiring as few as 40 training samples to achieve superior performance across the VSR, TallyQA, and GQA benchmarks. Finally, our comprehensive quantitative and qualitative analyses validate the impact of the super-resolution module in shifting the model toward agent-oriented behaviors, while highlighting the critical role of curriculum learning dynamics in optimizing both reasoning quality and overall performance.

%
%
\bibliographystyle{splncs04}
\bibliography{main}

\appendix
\section{Prompt Degisn Details}
This section details the prompt design for training Mags-RL. Specifically, we describe the prompts utilized during training in Sec.~\ref{subsec:prompt-train} and further discuss those employed for evaluation in Sec.~\ref{subsec:prompt-test}.

\subsection{Prompt Design for Training}\label{subsec:prompt-train}
As detailed in the main manuscript, training Mags-RL employs a Curriculum Learning (CL) framework divided into two distinct stages. In the first stage, the model is trained on 20 foundational samples to establish the core reasoning capabilities necessary for adhering to syntax requirements, correctly triggering external modules, and verifying previous logic; the corresponding system prompt and prompt suffix are illustrated in Fig.~\ref{prompt:system-prompt-1} and Fig.~\ref{prompt:suffix-1}, respectively. In the second CL stage, we train the model with additional 20 complex images,  the system prompt and prompt suffix are slightly modified to maintain consistency with the updated zoom-accuracy reward functions, which are designed to elicit specific behavioral shifts. Specifically, the primary modification to the system prompt occurs in Step 1, as shown in Fig.~\ref{prompt:system-2}, while the updated prompt suffix is outlined in Fig.~\ref{prompt:suffix-2}. Finally, the answer accuracy reward $r_{\text{ans}}$ incorporates reward signals from LLM judges to incentivize semantically aligned responses; we adopt the same prompt as in~\cite{fan2025grit} to facilitate this reward signaling.

\begin{promptbox}[label=prompt:system-prompt-1]{The system prompt for curriculum learning Stage 1. }
You are a precision visual reasoning agent. Your goal is to answer the user's question with the highest possible accuracy by using a zoom tool to verify details.\\

\textbf{Operational Protocol:}\\
You must follow this strictly sequential process:\\

\textbf{Step 1: Initial Reasoning \& Tool Trigger}\\
- Analyze the image and the question.\\
- Output your analysis inside <think>...</think> tags.\\
- If specific details are too small or unclear, you MUST use the zoom tool inside your thought block.\\
- \textbf{Zoom Format:} <zoom>[[x1, y1, x2, y2]]</zoom>\\
- Use double brackets for the coordinates.\\
- Coordinates are [top-left-x, top-left-y, bottom-right-x, bottom-right-y].\\
- \textbf{Constraint:} The box must be smaller than 40\% of the image area. Focus on the target, not the whole image.\\
- \textbf{Critical:} Stop generating immediately after closing the </think> tag.\\

\textbf{Step 2: System Execution (Automatic)}\\
- The system will process your zoom command.\\
- You will receive either high-resolution crops (if valid) or a failure message (if invalid).\\

\textbf{Step 3: Re-evaluation \& Conclusion}\\
- Once you receive the system feedback, analyze the new information (or the failure message).\\
- Output your updated reasoning inside <rethink>...</rethink> tags.\\
- Finally, provide the definitive answer inside <answer>...</answer> tags.\\

\textbf{Strict Formatting Rules:}\\
1. All initial analysis must be inside <think>...</think>.\\
2. The Zoom tool <zoom>...</zoom> must be nested INSIDE the <think>...</think> block.\\
3. All updated reasoning must be inside <rethink>...</rethink>.\\
4. The final answer must be inside <answer>...</answer>.
\end{promptbox}

\begin{promptbox}[label=prompt:suffix-1]{The prompt suffix in stage 1}
First, think between <think> and </think>, using <zoom>[[x1, y1, x2, y2]]</zoom> if details are unclear. Then, after receiving system feedback, provide your final reasoning in <rethink>... </rethink> and the final answer in <answer>...</answer>. \end{promptbox}

\begin{promptbox}[label=prompt:system-2]{Modified system prompt for curriculum learning Stage 2.}
Step 1: Initial Reasoning \& Tool Trigger\\
- Analyze the image and the question.\\
- Output your analysis inside <think>...</think> tags.\\
- You MUST use the zoom tool inside your thought block to verify details and count objects.\\
- \textbf{Zoom Format:} <zoom>[[x1, y1, x2, y2], [x3, y3, x4, y4], ...]</zoom>\\
- Use double brackets for the coordinates. You can output multiple boxes if there are multiple targets.\\
- Coordinates are [top-left-x, top-left-y, bottom-right-x, bottom-right-y].\\
- \textbf{Constraint:} Each individual box must be smaller than 40\% of the image area. Focus on specific targets, not the whole image.\\
- \textbf{Critical:} Stop generating immediately after closing the </think> tag.
\end{promptbox}

\begin{promptbox}[label=prompt:suffix-2]{The prompt suffix in stage 2}
First, think between <think> and </think>, using <zoom>[[x1, y1, x2, y2], ...]</zoom> to verify all relevant details. Then, after receiving system feedback, provide your final reasoning in <rethink>...</rethink> and the final answer in <answer>...</answer>.
\end{promptbox}

\subsection{Prompt Design for Evaluation}\label{subsec:prompt-test}
\begin{figure}[!t]
    \centering
    \includegraphics[width=0.95\textwidth]{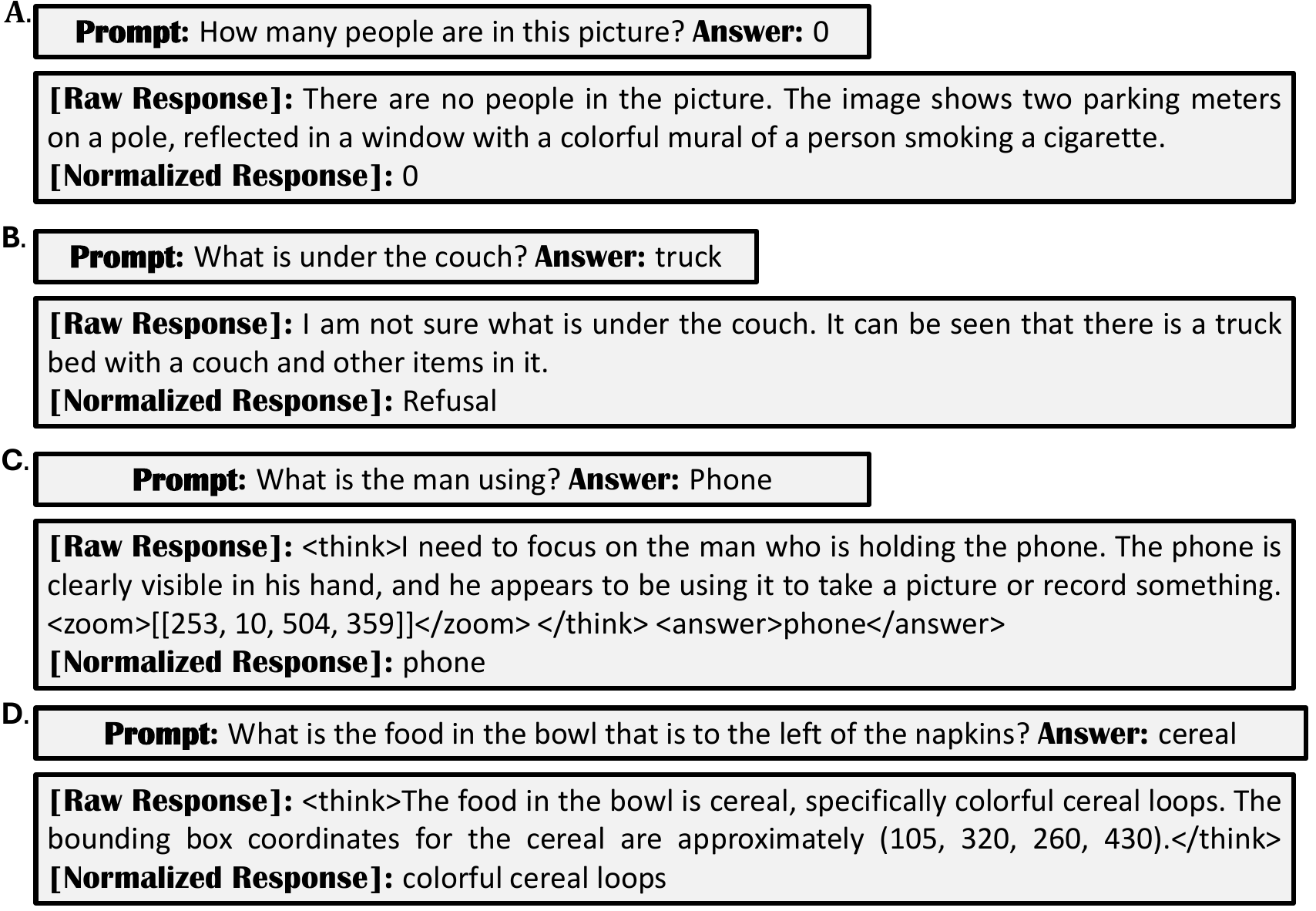}
    \caption{Illustration of the influence of data extractor module $E_{\text{data}}$, \textbf{A} and \textbf{B} represent response from direct query, \textbf{C} represent the response from ICL baseline and \textbf{D} denote the answer from GRIT. We omit the related images for clarity.}
    \label{fig:filter-influence}
\end{figure}

For our primary evaluation, we employ the system prompt and suffix from CL Stage 2 during model inference (see Fig.~\ref{prompt:system-2} and Fig.~\ref{prompt:suffix-2}). In contrast, the ablation study regarding the super-resolution module utilizes the prompt design from CL Stage 1. Furthermore, since evaluations across Zoom-Easy, Zoom-Medium, and Zoom-Hard involve comparing instructional models (e.g., direct query) with reasoning models (e.g., ours), we implement a strict data extractor module (i.e., $E_{\text{data}}$), to mitigate potential biases related to response length and verbosity. This module extracts key answers from all baselines to ensure a fair comparison; the prompt design and filtered results for this process are detailed in Fig.~\ref{prompt:data-filter} and Fig.~\ref{fig:filter-influence}, respectively. Notably, if the $E_{\text{data}}$ module fails to properly recognize a model's response (see Fig.~\ref{fig:filter-influence}\textbf{B}), it outputs "Refusal" as the final result, which is subsequently assigned a score of zero in the final performance calculation. Additionally, we observe that GRIT struggles to adhere to strict syntax requirements (see Fig.~\ref{fig:filter-influence}\textbf{D}); therefore, we apply the same filtering process used for the instructional models to maintain consistency. 

\begin{promptbox}[label=prompt:data-filter]{The prompt design for data extractor $E_{\text{data}}$}
\noindent \textbf{[System Prompt]:} You are a strict data-extraction assistant. Your only job is to extract the final, core factual answer from the model's response based on the original question.\\
- You must output ONLY the extracted concise answer.\\
- DO NOT include conversational filler (e.g., 'The image shows...', 'The answer is...').\\
- DO NOT use punctuation unless it is part of the answer itself.\\
- If the response implies the model cannot answer the question, output exactly: 'Refusal'.\\
\noindent \textbf{[User Prompt]:} \texttt{Question:} [query], \texttt{Model Response:} [raw text of model response], \texttt{Extracted Answer:} [final normalized response]
\end{promptbox}

After passing each model's response through the data extractor $E_{\text{data}}$, we utilize the resulting normalized, concise answers for our final evaluation. This assessment includes GPT accuracy, which evaluates semantic performance, and inclusion accuracy, which serves as a proxy for hard recall. The specific prompt design for the LLM judges is outlined in Fig.~\ref{prompt:gpt-evaluator}.

\begin{promptbox}[label=prompt:gpt-evaluator]{The prompt design for LLM judge}
\noindent \textbf{[System Prompt]:}You are an impartial, strict expert judge evaluating the factual correctness of a model's answer to a question, based solely on the provided question and ground-truth answer.\\
You must score the model's answer on a scale from 0.0 to 1.0 using the following strict rubric:\\
- 1.0: The answer is factually correct, complete, and perfectly aligns with the ground truth.\\
- 0.75: The answer is mostly correct and relevant but is missing a very minor detail.\\
- 0.5: The answer is partially correct but misses major parts of the ground truth or includes some irrelevant info.\\
- 0.25: The answer is mostly incorrect but contains a tiny sliver of relevant truth.\\
- 0.0: The answer is completely incorrect, irrelevant, or contradicts the ground truth.\\
RULES:\\
- DO NOT give preference to conversational, wordy, or detailed responses.\\
- A concise, short answer MUST receive a 1.0 if it captures the core facts of the ground truth.\\
- DO NOT penalize for grammatical incompleteness.\\
- Output ONLY the float value (e.g., 0.0, 0.25, 0.5, 0.75, or 1.0). Do not include any other text.\\
\noindent \textbf{[User Prompt]:} Question: [question], Ground Truth: [ground\_truth], Model Answer: [normalized prediction], Score: [final score]
\end{promptbox}
Finally, our evaluation benchmarks (i.e., the Zoom-Easy, Zoom-Medium and Zoom-Hard) are constructed using a VLM-as-a-judge framework to categorize samples by their complexity levels. The specific selection prompt utilized for this process is outlined in Fig.~\ref{prompt:qwen-selection}.

\begin{promptbox}[label=prompt:qwen-selection]{The prompt design for VLM judge}
\noindent \textbf{[System Prompt]:} You are an expert Visual Information Analyst. Your task is to evaluate the "Information Density" and "Zoom Necessity" of the provided image.\\
Analyze the image based on these criteria:\\
1. Object Scale: How small are the key elements relative to the image size?\\
2. Visual Clutter: Is the scene crowded, chaotic, or clean?\\
3. Text/Detail Level: Is there fine print, tiny textures, or distant background details that are hard to see?\\
Based on your analysis, provide a "zoom\_score" from 1 to 10:\\
- Score 1-3 (Simple): \\
- Subject is large, centered, and clearly visible.\\
- No zoom needed. \\

- Score 4-7 (Medium): \\
- A standard scene with multiple objects or moderate distance.\\ 
- Main elements are visible, but background details might be blurry. \\
- Zooming would help clarify relationships but isn't strictly mandatory for the gist.\\

- Score 8-10 (Hard / Complex): \\
- High Zoom Necessity. The image contains tiny, critical details.\\
- Without zooming, it is impossible to distinguish individual elements.\\

Output Format (JSON only):\\
\texttt{\{ reasoning: [reasoning content], zoom\_score: [complexity score] \}}
\end{promptbox}

\section{Reward Function Details.}
This section details the reward function design for training Mags-RL. Specifically, the training process involves three primary reward signals: the format reward $r_{\text{fmt}}$, the answer accuracy reward $r_{\text{ans}}$, and the zoom accuracy reward $r_{\text{zoom}}$. Together, these encourage the model to adhere to syntax constraints, provide correct and concise answers, and properly trigger the external visual module $S_a$. Additionally, an optional, length-based rethink volume reward $r_{\text{revo}}$ is incorporated to encourage the generation of substantive rethinking content. We detail these reward definitions below.

\noindent \textbf{Format Reward $r_{\text{fmt}}$}. This reward incentivizes the policy model $\pi_\theta$ to generate output sequences $o$ that adhere to strict formatting constraints. Specifically, $r_{\text{fmt}} := r_{\text{afmt}} + r_{\text{tfmt}} + r_{\text{rfmt}}+ r_{\text{zfmt}}$, where $r_{\text{afmt}}, r_{\text{tfmt}}$, $r_{\text{rfmt}}$ denote the format rewards for the answer, thinking, and rethinking phases, respectively. For example, we assign $r_{\text{afmt}} = 1.0$ if and only if the \texttt{<answer>} and \texttt{</answer>} tags appear in $a$ in the correct sequential order. Furthermore, $r_{\text{tfmt}}$ and $r_{\text{rfmt}}$ are each capped at a maximum value of 0.5. 

The zoom format reward $r_{\text{zfmt}}$ distinguishes itself from the other three format signals by simultaneously enforcing syntax requirements and facilitating the agent-triggering mechanism. Formally, this reward is defined as: $r_{\text{zfmt}} := I_{\text{zfmt}} \cdot I_{N_u(\mathbf{o}_{\text{zoom}}^{(1)})\geq5}\cdot [0.1 * I_{f_d(\mathbf{o}_{\text{zoom}}^{(1)})\leq 0.4} + I_{f_d(\mathbf{o}_{\text{zoom}}^{(1)})\geq 0.4} \cdot (0.5 + 0.5 * \min (1.0, \log(N_u(\mathbf{o}_{\text{zoom}}^{(1)})+1) / \log20))]$. Here, $I_{\text{zfmt}}$ acts as a gated binary indicator to ensure format consistency, while $I_{N_u(\mathbf{o}_{\text{zoom}}^{(1)})\geq 5}$ serves as a safeguard against empty or trivial reasoning steps. Furthermore, $I_{f_d(\mathbf{o}_{\text{zoom}}^{(1)})\leq 0.4}$ is a binary indicator derived from the lexical diversity function $f_d(\cdot)$ (i.e., defined as the ratio of unique words to the total reasoning length). This guarantees that zooming operations are only highly rewarded when triggered by diverse, substantive reasoning. Finally, the $\min(\cdot,\cdot)$ clipping and $\log(\cdot)$ scaling are incorporated to prevent reward hacking, effectively neutralizing the advantage of excessively verbose outputs. Because the maximum value for $r_{\text{zfmt}}$ is capped at 1.0, the total maximum format reward $r_{\text{fmt}}$ sums to 3.0.

\noindent \textbf{Answer Accuracy Reward $r_{\text{ans}}$}. This reward serves as the primary signal to ensure answer accuracy. Because we expect a reasoning model to confine its logical deductions to the intermediate thinking components and ultimately produce a concise final response, we formulate the reward as $r_{\text{ans}} := I_{\text{afmt}}\cdot \max(I_{a == a^*}, 0.5 \cdot I_{\text{gpt score}\geq 0.7})$. Here, $I_{a == a^*}$ represents a binary indicator function that evaluates whether the model's response $a$ exactly matches the ground-truth answer $a^*$. Furthermore, $I_{\text{gpt score}\geq 0.7}$ acts as a secondary indicator function evaluating whether the GPT-assigned score for the $(q,a^*,a)$ triplet is 0.7 or higher. By providing this tiered accuracy signal, we incentivize the model to generate succinct responses (rewarding exact matches maximally) while still offering partial credit for semantically aligned answers via the GPT score (e.g., "cat" vs. "kitty"). 

\noindent \textbf{Zoom Accuracy Reward $r_{\text{zoom}}$}. Training the policy model $\pi_\theta$ to localize the appropriate regions within $I$ to effectively answer the query $q$ without any ground-truth bounding box or external grounding tools is not trivial. Consequently, we employ a curriculum learning approach, which yields distinct reward signals for the two training stages, denoted as $r_{\text{zoom}_1}$ and $r_{\text{zoom}_2}$. The reward function is formulated as follows:
\begin{equation}
    r_{\text{zoom}_i} := \begin{cases}
        T \cdot f_{\text{box}}(Z), & \text{if stage } i = 1 \\
        S \cdot \max(0, \min(1.0, h_{\text{box}}(Z)) - 0.05 \cdot g_{\text{box}}(Z)), & \text{if } i = 2, \text{counting} \\
        S \cdot f_{\text{box}}(Z), & \text{if } i = 2, \text{others}
    \end{cases}
\end{equation}$
\text{where } \left\{
\begin{aligned}
    & T := I_{\text{zfmt}} \cdot I_{N_u(\mathbf{o}_{\text{zoom}}^{(1)})\geq 5} \cdot [0.1 \cdot I_{f_d(\mathbf{o}_{\text{zoom}}^{(1)})\leq 0.4} + I_{f_d(\mathbf{o}_{\text{zoom}}^{(1)})\geq 0.4}] \\
    & S := I_{\text{zfmt}} \cdot (0.1 + 0.9 \cdot I_{N_u(\mathbf{o}_{\text{zoom}}^{(1)})\geq 5} \cdot I_{f_d(\mathbf{o}_{\text{zoom}}^{(1)})\geq 0.4}) \\
    & f_{\text{box}}(Z) := k/n \\
    & h_{\text{box}}(Z) := k / |a^*| \\
    & g_{\text{box}}(Z) := n-k
\end{aligned} \right.$

\noindent Here, $k,n$ represent the number of valid and total bounding box predictions, respectively. $f_{\text{box}}(Z)$ represents the precision of the bounding box predictions, while $h_{\text{box}}(Z)$ denotes the recall relative to the ground-truth object count $|a^*|$. The penalty term $g_{\text{box}}(Z)$ computes the number of incorrect bounding box predictions, effectively preventing reward hacking where $\pi_\theta$ might exploit the system by generating excessive random guesses. The valid bounding boxes number m is determined by rule check (e.g., validation of coordinates and size limitation of cropped area). The scaling factors $S$ and $T$ ensure that valid bounding box predictions are strictly grounded in diverse and meaningful reasoning content $w_1$. 

During stage-1 training on simpler data, $r_{\text{zoom}_1}$ incentivizes $\pi_\theta$ to make accurate but conservative predictions (e.g., favoring a single, safe bounding box per query). In stage-2 training, by explicitly shifting to a recall-based reward for counting questions (e.g., using TallyQA data), the model is compelled to make more aggressive predictions, extracting all necessary visual components. Finally, this reward function is also capped at a maximum value of 1.0.

\noindent \textbf{Rethink Volume Reward $r_{\text{revo}}$}. This optional reward is designed to help the policy model $\pi_\theta$ to generate substantial and meaningful rethinking content $\mathbf{o}_{\text{think}}^{(2)}$. It is formulated as:
$r_{\text{revo}} := (0.5 + 0.5 * I_{\text{afmt}}) \cdot I_{N_u(\mathbf{o}_{\text{think}}^{(2)})\geq 5}\cdot \min(1.0, 0.2 * \sqrt{N_u(\mathbf{o}_{\text{think}}^{(2)})})$. Here, $I_{\text{afmt}}$ is a binary indicator variable specifying whether the answer $a$ adheres to the format constraints, which ensures the model is only fully incentivized to rethink if it successfully produces a properly formatted response. Furthermore, $I_{N_u(\mathbf{o}_{\text{think}}^{(2)})\geq 5}$ acts as an indicator function that verifies the rethinking content $\mathbf{o}_{\text{think}}^{(2)}$ contains a minimum of 5 unique words, with $N_u(\cdot)$ representing a regex-based function that calculates the unique word count. Finally, the $\min(\cdot,\cdot)$ clipping function prevents reward hacking, ensuring the model is not disproportionately rewarded for excessively verbose outputs. This reward function is capped at a maximum value of 1.0.

The final group-normalized advantage $\hat{A}_i$ is calculated based on the total reward signal $r_{\text{total}}$, which is defined as:
\[r_{\text{total}} := \lambda_1\cdot r_{\text{fmt}} + \lambda_2 \cdot r_{\text{ans}} + \lambda_3 \cdot r_{\text{zoom}} + \lambda_4 \cdot r_{\text{revo}} \]
\noindent Here, the weighting coefficients are set to $\lambda_2 = 2.0$, $\lambda_3 = 1.0$, and $\lambda_4 = 0.5$. For the format reward represented by $\lambda_1$, we apply varying weights to its underlying components: the zoom format component ($r_{\text{zfmt}}$) receives a weight of 0.5, while the remaining formatting sub-components each receive a weight of 0.1.

\section{Implementation Details}
\noindent \textbf{Training Implementation Details}. We employ Qwen2.5-VL-3B as our base model without using any cold-start initialization. We conduct experiments across two CL stages using the AdamW optimizer with a global batch size of 64 and a GRPO group size of $G = 16$. Additionally, we integrate the SMFANet~\cite{smfanet} as our super-resolution module. In stage 1, we train for 300 steps with a learning rate of $2\times 10^{-6}$, a sampling temperature of 0.09, and a KL penalty coefficient of $\beta = 0.04$, using the base model as the reference policy (i.e., $\pi_{\text{ref}}$). In stage 2, we restart for another 225 steps with a reduced learning rate of $5\times 10^{-7}$, an increased temperature of 1.0, and $\beta = 0.03$, updating $\pi_{\text{ref}}$ to the weights obtained from Stage 1. All training is performed on $4\times$ NVIDIA A100 (80GB) GPUs with DeepSpeed ZeRO-2~\cite{rasley2020deepspeed} optimization. Additionally, we use GPT-4 (i.e., gpt-4o-mini) as our LLM-judges to provide score for building answer accuracy reward $r_{\text{ans}}$. 

\noindent \textbf{Testing Implementation Details.} Throughout our experiments, we utilize subsets from the VSR~\cite{liu2023visual}, TallyQA~\cite{acharya2019tallyqa}, and GQA~\cite{hudson2019gqa} benchmarks to comprehensively evaluate the model's visual understanding. Specifically, VSR assesses spatial relationship comprehension, TallyQA evaluates fine-grained instance perception and object individuation through counting, and GQA tests the model's capacity for compositional reasoning and visual grounding. To rigorously assess performance across varying levels of difficulty, we stratify the evaluation data into three disjoint subsets: \texttt{Zoom-Easy}, \texttt{Zoom-Medium}, and \texttt{Zoom-Hard}. As discussed previously, we employ a VLM-as-a-judge framework (specifically, \texttt{Qwen3-VL-4B-Instruct}) to determine the complexity of each sample based on the ground-truth query and the corresponding image (see Fig.~\ref{prompt:qwen-selection} for the selection prompt). After filtering out unclear query-answer pairs, the final sample counts for each dataset across the three complexity categories are detailed in Table~\ref{tab:data-simple}.

\begin{table*}[ht]
\centering
\caption{Dataset distribution across the Zoom-Easy, Zoom-Medium, and Zoom-Hard complexity categories.}
\label{tab:data-simple}
\scriptsize
\renewcommand{\arraystretch}{1.15} 
\setlength{\tabcolsep}{4pt}

\begin{tabular*}{0.8\textwidth}{@{\extracolsep{\fill}} l c c c @{}}
\toprule
\rowcolor{tablegrey} 
\textbf{Complexity Category} & \textbf{VSR} & \textbf{TallyQA} & \textbf{GQA} \\
\midrule
Zoom-Easy   & 288  & 491 & 500 \\
Zoom-Medium & 1000 & 500 & 999 \\
Zoom-Hard   & 46   & 139 & 365 \\
\bottomrule
\end{tabular*}
\end{table*}
\noindent During evaluation, we employ a near-greedy decoding strategy by setting the temperature to $0.01$ and the top-$p$ parameter to $0.95$. Additionally, when calculating the GPT accuracy score, we utilize the \texttt{gpt-4o-mini} model as our LLM judge (see Fig.~\ref{prompt:gpt-evaluator} for the specific prompt design).

\section{Case Studies}
\begin{figure}[!t]
    \centering
    \includegraphics[width=\textwidth]{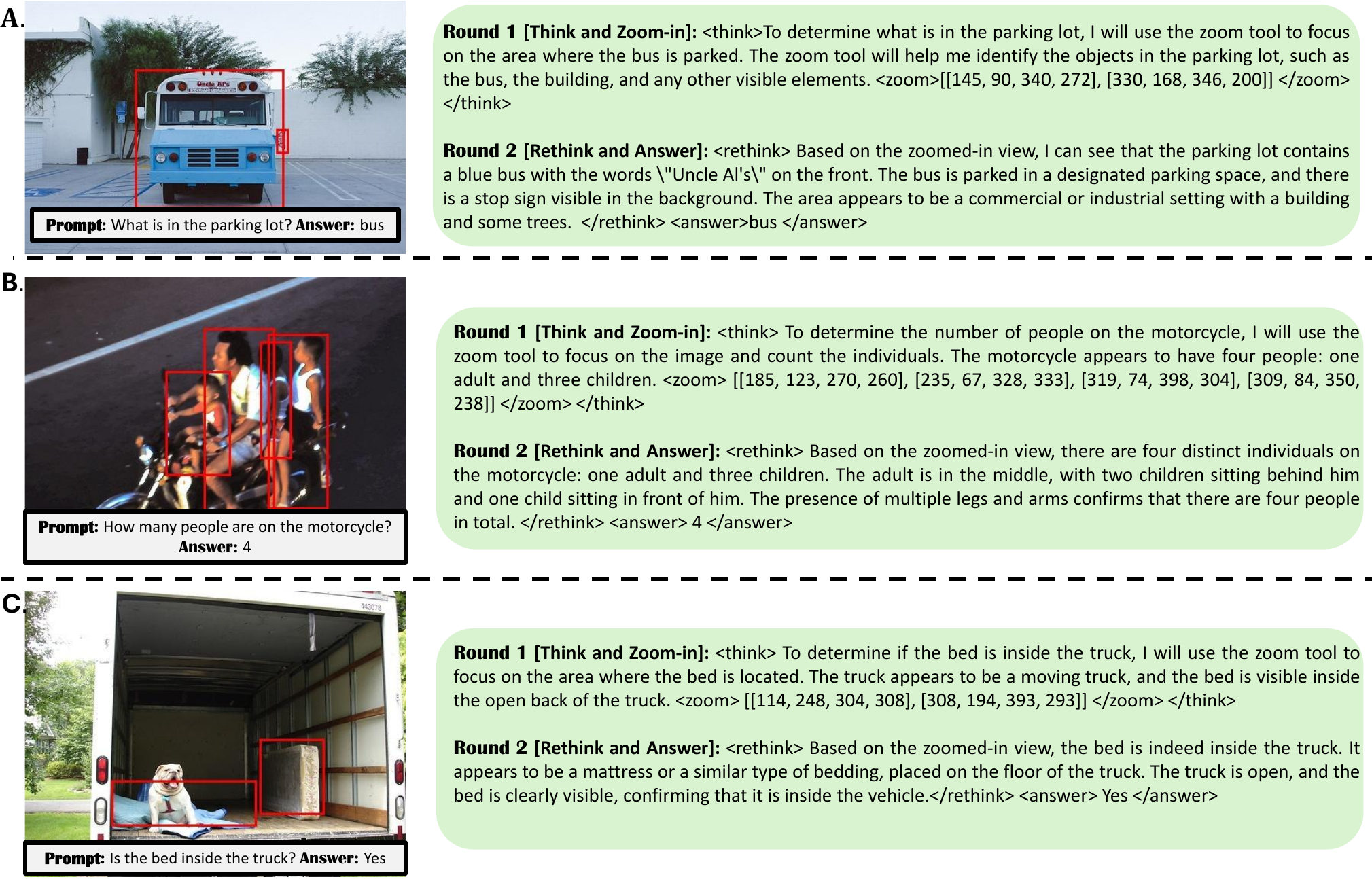}
    \caption{Illustration of Mags-RL responses on the Zoom-Easy evaluation set. The red boxes indicate the spatial regions explicitly referenced in the model's text response. }
    \label{fig:example-final-easy}
\end{figure}

\begin{figure}[!t]
    \centering
    \includegraphics[width=\textwidth]{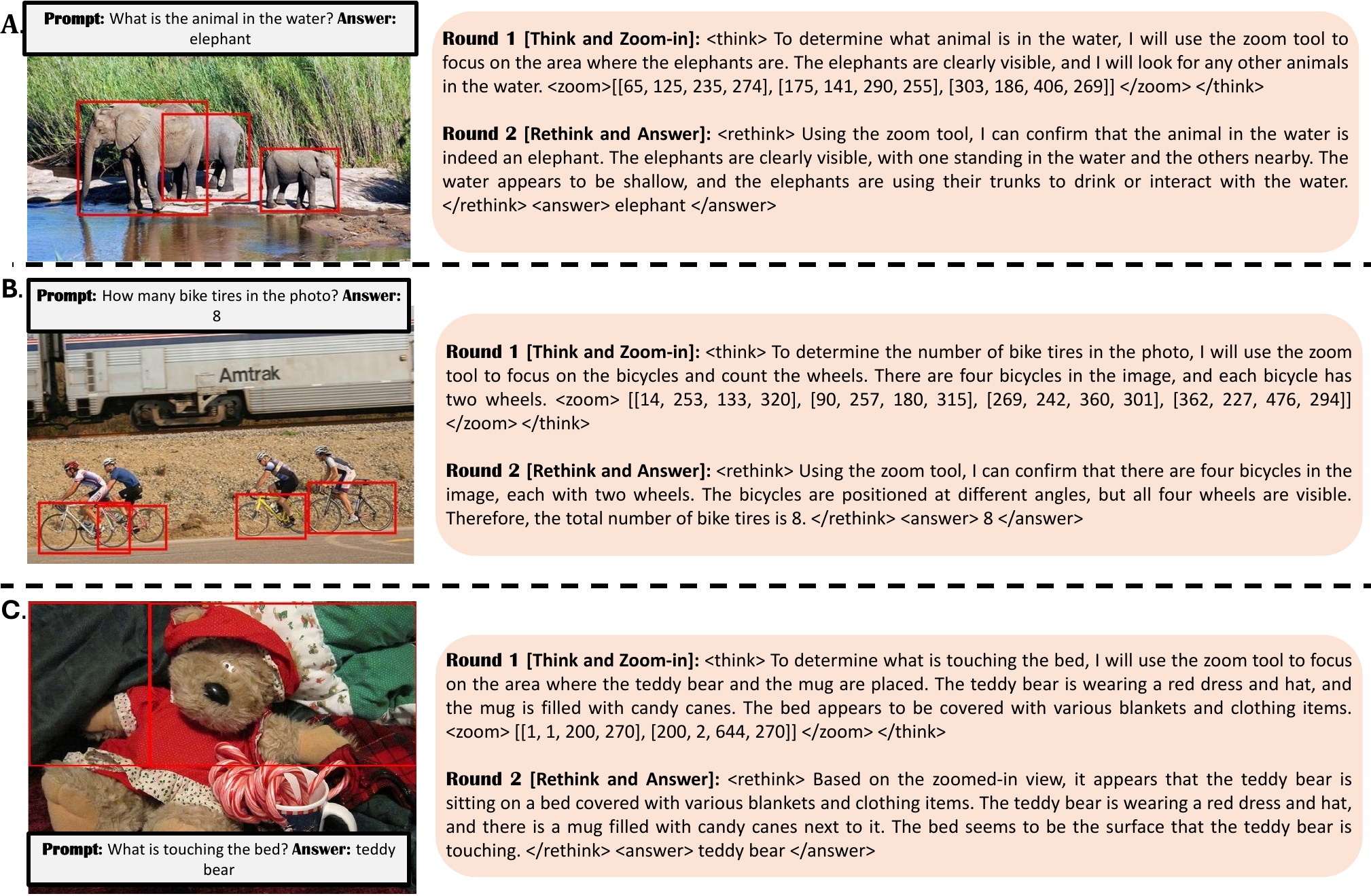}
    \caption{Illustration of Mags-RL responses on the Zoom-Medium evaluation set. The red boxes indicate the spatial regions explicitly referenced in the model's text response. }
    \label{fig:example-final-medium}
\end{figure}

\begin{figure}[!t]
    \centering
    \includegraphics[width=\textwidth]{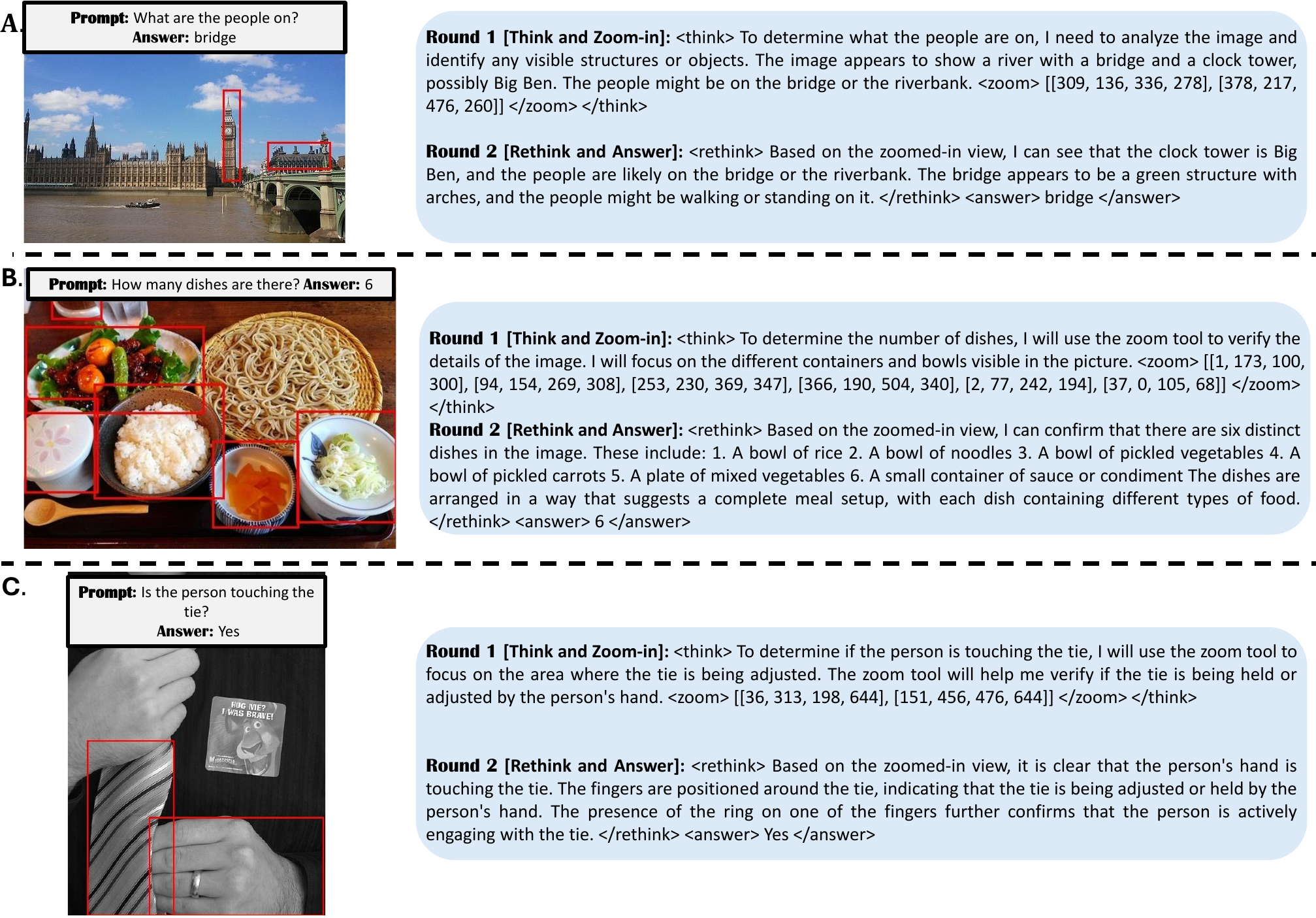}
    \caption{Illustration of Mags-RL responses on the Zoom-Hard evaluation set. The red boxes indicate the spatial regions explicitly referenced in the model's text response. }
    \label{fig:example-final-hard}
\end{figure}

We present additional case studies in Fig.~\ref{fig:example-final-easy}, Fig.~\ref{fig:example-final-medium} and Fig.~\ref{fig:example-final-hard}, where all responses are generated by Mags-RL. Specifically, these examples demonstrate that our method enables the model to accurately identify relevant bounding boxes whenever they are beneficial for visual comprehension. Notably, our framework achieves this without relying on any external grounding networks.


\end{document}